\documentclass{article}

\usepackage[preprint]{corl_2022} 
\usepackage{amsmath}
\usepackage{amsfonts}
\usepackage{cleveref}
\usepackage{subcaption}
\usepackage{todonotes}
\usepackage{wrapfig}
\usepackage{booktabs}
\usepackage{multicol}
\usepackage{xcolor}

\title{Learning Dense Visual Descriptors using Image Augmentations for Robot Manipulation Tasks}

\author{
	Christian Graf$^{\dagger,1}$, 
	David B. Adrian$^{\dagger,1,2}$, 
	Joshua Weil$^{1,3}$, 
	Miroslav Gabriel$^1$, \\ \\
	\textbf{	Philipp Schillinger$^1$, Markus Spies$^1$, 
		Heiko Neumann$^2$, 
		Andras Kupcsik$^1$} \\ \\
	$^\dagger$Equal contribution, $^1$Bosch Center for Artifical Intelligence,\\ $^2$Ulm University, $^3$KTH Royal Institute of Technology
}

\newcommand{\dist}[1]{\mbox{dist}\left(#1\right)}

\begin{document}
\maketitle


\begin{abstract}
We propose a self-supervised training approach for learning view-invariant dense visual descriptors using image augmentations.
Unlike existing works, which often require complex datasets, such as registered RGBD sequences, we train on an unordered set of RGB images.
This allows for learning from a single camera view, e.g., in an existing robotic cell with a fix-mounted camera.
We create synthetic views and dense pixel correspondences using data augmentations.
We find our descriptors are competitive to the existing methods, despite the simpler data recording and setup requirements.
We show that training on synthetic
correspondences provides descriptor consistency across a broad range of camera views.
We compare against training with geometric correspondence from multiple views and provide ablation studies.
We also show a robotic bin-picking experiment using descriptors learned from a fix-mounted camera for defining grasp preferences.

\end{abstract}



\section{Introduction}

\begin{wrapfigure}{r}{.28\textwidth}
	\vspace{-6mm}
	\begin{center}
		\includegraphics[width=.29\textwidth]{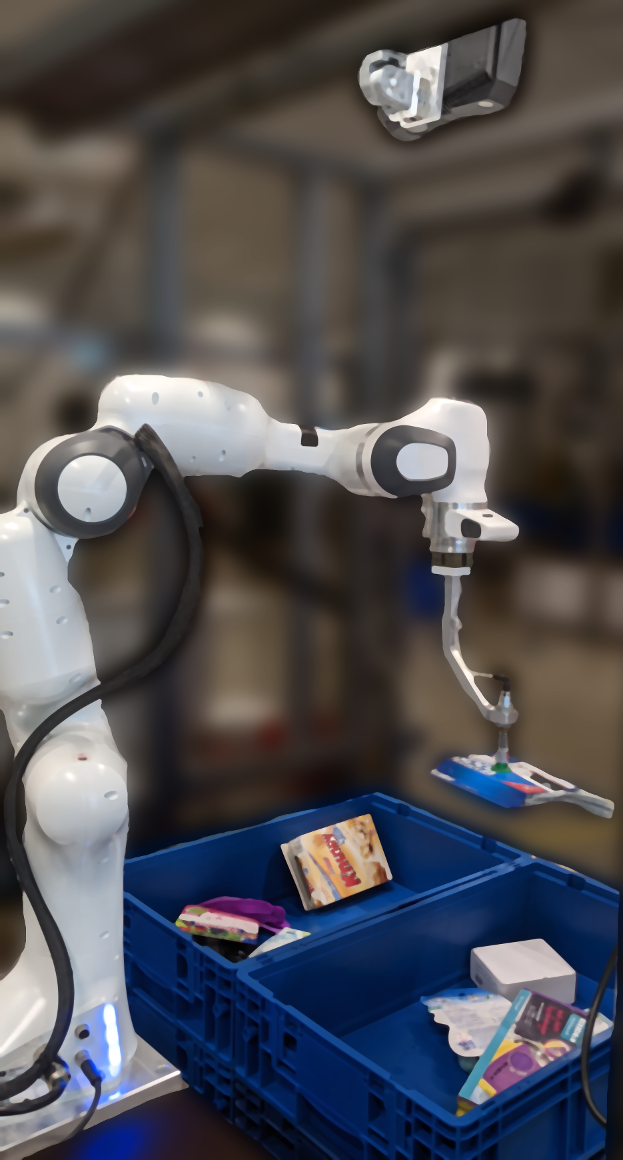}
	\end{center}
	\caption{Bin-picking setup with a single, fix-mounted overhead camera.}
	\label{fig:bin_picking_setup}
	\vspace{-10mm}
\end{wrapfigure}

Scene and object understanding is essential for robot manipulation tasks, including assembly or bin picking. 
Often, the representation of choice is task-specific segmentation or pose estimation, trained in a supervised manner with labeled data. 
Labeling, however, is expensive and time-consuming, which is why self-supervised learning of dense visual descriptors has recently gained substantial attention in the robotics community, inspired by the works of \citet{Schmidt2017} and \citet{Florence2018}.

Dense Object Nets (DONs) proposed by  \citet{Florence2018} learn dense visual descriptors of objects fully self-supervised in a robotic environment. 
The learned descriptors are view-invariant, show potential for within-class generalization and they naturally apply to non-rigid objects.
The dense descriptor representation can be flexibly used for various downstream robotic tasks, such as, grasping (\citet{Florence2018, Kupcsik2021, Adrian2022}), rope manipulation (\citet{Sundaresan2020}) and learning control (\citet{Manuelli2020, Florence2020}).

Self-supervised training of DONs, however, relies on pixel correspondences across multiple camera views provided by a registered RGBD image sequence, which requires accurate camera calibration and pose recording.
Furthermore, pixel correspondence tends to be inaccurate with inexpensive depth cameras, even with data preprocessing.
Finally, data collection is constrained by robot kinematics and the need for an expert setting up and supervising the procedure.

In this paper we relax these assumptions fundamentally and instead of a complex setup with a single, robot-mounted moving camera, or multiple static ones, we solely rely on an unordered set of RGB images to learn object descriptors, for example, recorded by a single fixed camera.
In our work, instead of relying on multi-view, \emph{geometric correspondence}, we use augmentations of single images to obtain alternative views and \emph{synthetic correspondence}.
This idea was already explored in computer vision by \citet{Thewlis2017} and \citet{Novotny2018}, in the context of learning geometrically consistent pixel-level descriptors across multiple object classes.
In this paper we show that relying on synthetic image augmentations achieves competitive performance in terms of keypoint tracking accuracy compared to a network trained with geometric correspondence.
Importantly, this approach can easily be adopted to existing industrial setups with fix-mounted cameras, or with cameras too heavy to be mounted on a robotic arm, without additional engineering effort.
We show such a robotic bin-picking setup in Fig.~\ref{fig:bin_picking_setup}, with an overhead, fix-mounted camera.

Our contributions are as follows: (i) we adapt existing work on training self-supervised pixel embeddings (\citet{Novotny2018}, \citet{Chen2020}) to robotic grasping downstream tasks. (ii) we show that for robotic grasping tasks our approach is en par with state-of-the-art (\citet{Adrian2022}) in terms of keypoint tracking accuracy while drastically simplifying the data collection, and finally (iii) we show a real-world robotic bin-picking experiment where human preference on grasp configuration is encoded with dense visual descriptors, with the constraint of using a single, fixed-mounted camera.


\section{Related work}
\label{sec:related_work}

In the following, we review recent work on self-supervised dense visual descriptor learning for robotic manipulation in more detail.
We also discuss related work on self-supervised representation learning and learning from a set of single images, which are core concepts in our work.

\textbf{Dense visual descriptors in robotic manipulation.}
Inspired by the work of \citet{Schmidt2017}, \citet{Florence2018} proposed self-supervised training of dense visual descriptors by and for robotic manipulation.
Their approach was later adopted by \citet{Florence2020} to learn from multi-view correspondence in dynamic scenes and showed an application for policy learning.
\citet{Sundaresan2020} applied the descriptor space representation to learn challenging rope manipulation in simulations.
Applying the dense descriptor representation to learn model predictive controllers was shown by \citet{Manuelli2020}. 
\citet{Vecerik2020} use multi-view consistency for keypoint detection and show an application for reinforcement learning. 

Another line of work investigated improved training strategies of dense visual descriptors. 
There are multiple contributions focusing on learning multi-object and multi-class descriptors by \citet{Yang2021, Hadjivelichkov2021} and \citet{Adrian2022}.
The work by \citet{Kupcsik2021} exploits known object geometry to compute optimal descriptor embeddings.
Finally, \citet{Yen-chen22} utilize NeRF to generate dense correspondence datasets from RGB images.
This alleviates problems with noisy depth data and proves especially helpful for thin and reflective objects.

Several papers proposed to learn directly from synthetic images composed of random backgrounds and randomly sampled, masked objects distributed over the image, see \citet{Florence2018, Chai2019, Yang2021}.
Learning from such synthetic images can be more efficient due to higher object density and ground truth correspondence, however, they rely on labeled datasets with object masks.
Masking is either achieved by 3D reconstruction with a robot wrist mounted camera (\citet{Florence2018, Chai2019}), or a labelled RGBD dataset (\citet{Yang2021}).
As opposed to image composition via mask-labeled datasets, the image augmentation technique, as in this paper, only requires an unordered, unlabelled RGB dataset. 
This significantly simplifies data collection and opens up the possibilities to learn dense visual descriptors where no 3D reconstruction is possible, or where object masks are not available.

\textbf{Self-supervised descriptor learning from RGB images}.
An intuitive way to generate geometric correspondence is to estimate the optical flow of subsequent frames from a video.
\citet{Deekshith2020} adopted this technique to train DONs using contrastive learning.
\citet{Thewlis2017} proposed to use optical flow from videos, or image augmentations to embed pixels of objects in view-invariant coordinate frames.
\citet{Novotny2018} adopted this method for pretraining of geometry-oriented tasks, such as object specific part detection in images.
\citet{Zhang2020} propose to learn pixel-wise descriptors from single images with augmentations by using hierarchical visual grouping of image patches based on contour.
Our work follows the image augmentation technique of single RGB images to generate alternative views and synthetic correspondence.
Equivariant network architectures such as proposed by~\citet{Cohen2016, Wang2021} could replace certain augmentations (e.g. rotation) during training and improve sample efficiency.
In our setup we can easily apply affine transformation on training data and we rely on a vanilla ResNet architecture for our experiments, which achieves good $SE(2)$ equivariance, as shown in our experiments.

\textbf{Self-supervised visual representations learning.}
Instead of training on large supervised datasets, self-supervised methods have become a popular way to obtain visual representations, which can be fine-tuned to specific downstream tasks.
A recent and very successful approach using contrastive learning is SimCLR by \citet{Chen2020}.
It aims to maximize agreement between two augmented versions of the same image, while considering all other images in the batch as negative samples.
\citet{Grill2020} proposed BYOL (Bootstrap your own latent) that, in comparison to contrastive methods, does not rely on the sampling of negatives.
With Barlow Twins \citet{Zbontar2021} also forgo negative samples by optimizing the cross-correlation matrix between embeddings from two augmented versions of the same image to be close to identity.
Our approach is most similar to SimCLR as we employ the same loss formulation, but with the important difference that our batch is constituted by individual pixel descriptors instead of full image embeddings.

\section{Method}
\label{sec:method}

\begin{figure}[h]
	\centering
	\includegraphics[width=.9\textwidth]{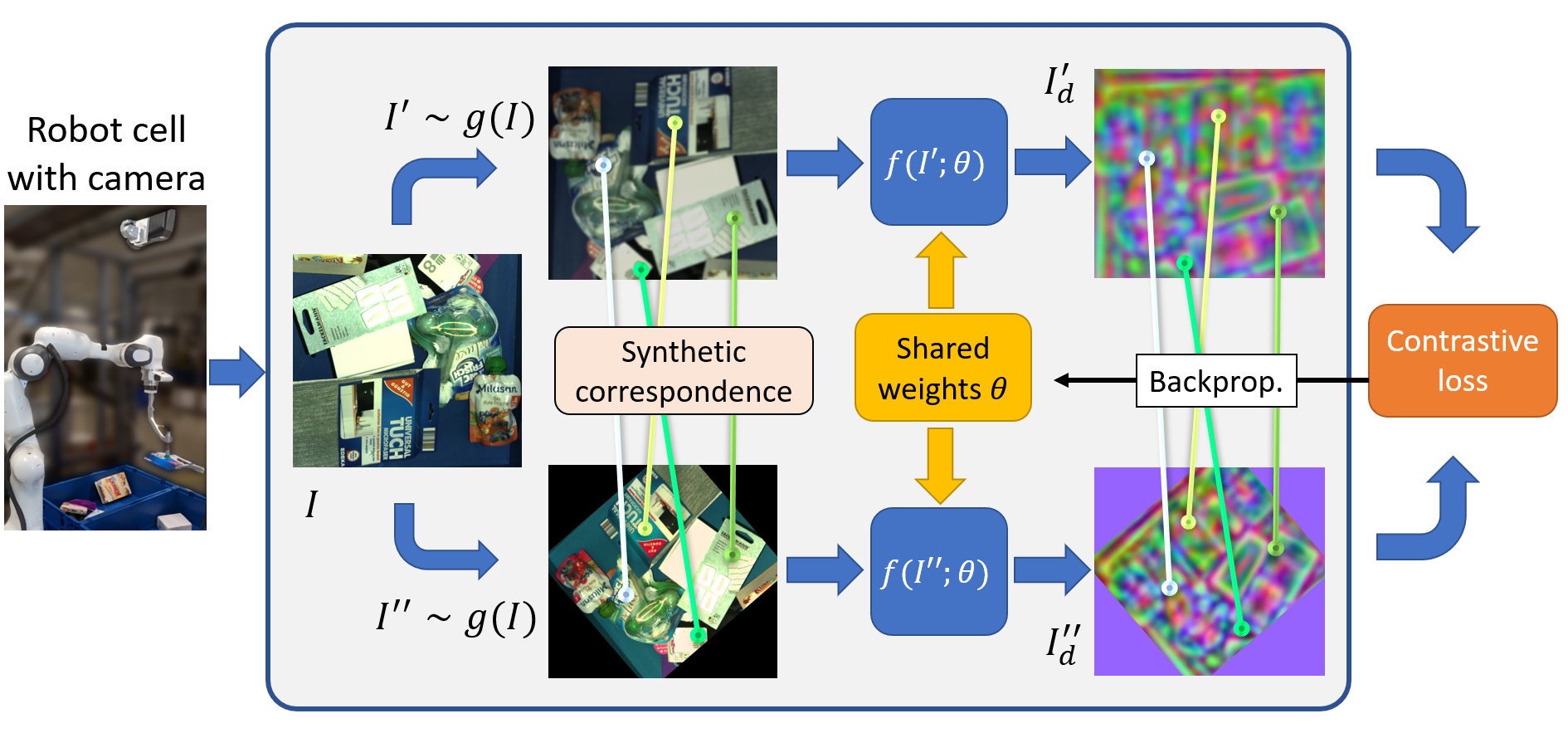}
	\caption{Illustration of the training pipeline. 
		For every RGB image $I$ in a batch, we sample a pair of augmented images $I'$ and $I''$ while keeping track of the pixel correspondences. 
		Then, we evaluate the RGB images with the trained fully-convolutional network $f(\cdot;\theta)$ with trainable parameters $\theta$. 
		Finally, we compute the contrastive loss and backpropagate the error using the embedded images $I'_d$ and $I''_d$ along with the correspondence information.}
	\label{fig:training_pipeline}
\end{figure}

In this section we discuss our proposed training approach using image augmentations.
We first give an overview of the whole training pipeline, then discuss image augmentation techniques and finally present the loss formulation and dataset requirements.
For an illustration of the training pipeline we refer to Fig.~\ref{fig:training_pipeline}.

\subsection{Dense Descriptor Training with Synthetic Correspondence}
\label{subsec:single_view_training}
Inspired by the work of \citet{Novotny2018} we rely on training on an unordered set of images and use image augmentations to arrive at alternative views of each image.
First, we sample a minibatch of $N$ RGB images from the training data set consisting of independent RGB images.  
For every image $I$ in the minibatch we sample two augmented views $I'\sim g(I)$ and $I''\sim g(I)$ by applying randomized augmentations $g:\mathbb{R}^{H\times W\times 3}\mapsto \mathbb{R}^{H\times W\times 3}$ (described in more detail below).
A learned  fully-convolutional network \cite{Long2017} model $f(\cdot;\theta),~f:\mathbb{R}^{H\times W\times 3}\mapsto \mathbb{R}^{H\times W\times D}$ maps the augmented images $I'$ and $I''$ to their descriptor space embeddings $I'_d$ and $I''_d$.
The user defined parameter $D\in \mathbb{N}^+$ controls the resolution of the descriptor space.

By keeping track of the position of each pixel in the original image $I$ during the augmentations, we sample pairs of pixel locations between $I'$ and $I''$ that share the same position in $I$.
We refer to these as \emph{synthetic} correspondences, emphasizing the use of synthetic image augmentations as opposed to \emph{geometric} correspondences coming from the 3D geometry of multiple camera views, as in \cite{Schmidt2017, Florence2018}.
The descriptor values at the sampled pixel correspondence locations serve as positive pairs for contrastive learning.

\subsection{Image Augmentations}
\label{sec:augmentations}
For robotic applications we require descriptors that are invariant to translations, rotations and perspective changes of the objects, as well as changes of lighting conditions.
In vanilla DONs training, cf. \citet{Florence2018}, this is achieved by recording a diverse training set of registered RGBD image sequences, which contain sufficient variance in camera and object poses, and to some extent in lighting conditions.
In our work, we achieve a similar effect purely by imposing data augmentations on single RGB images from an unordered set.

\begin{figure}
	\centering
	\begin{subfigure}[b]{0.16\textwidth}
		\centering
		\includegraphics[width=\textwidth]{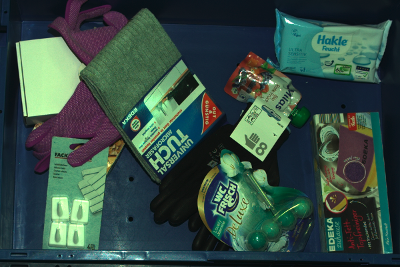}
		\caption{Original}
	\end{subfigure}
	\hfill
	\begin{subfigure}[b]{0.16\textwidth}
		\centering
		\includegraphics[width=\textwidth]{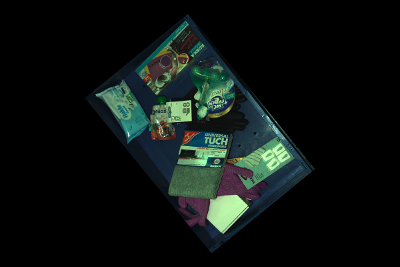}
		\caption{Affine}
	\end{subfigure}
	\hfill
	\begin{subfigure}[b]{0.16\textwidth}
		\centering
		\includegraphics[width=\textwidth]{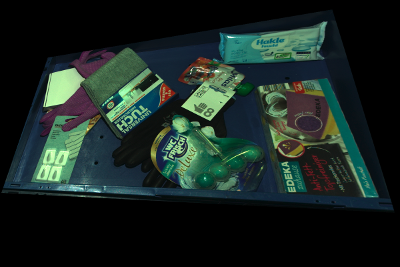}
		\caption{Perspective}
	\end{subfigure}
		\hfill
	\begin{subfigure}[b]{0.16\textwidth}
		\centering
		\includegraphics[width=\textwidth]{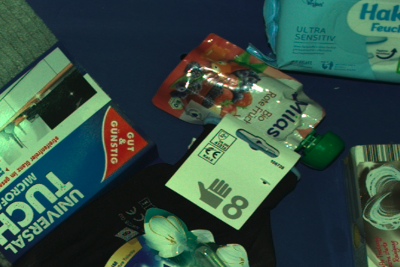}
		\caption{Resize\&Crop}
	\end{subfigure}
	\hfill
	\begin{subfigure}[b]{0.16\textwidth}
		\centering
		\includegraphics[width=\textwidth]{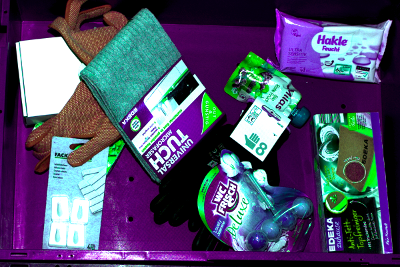}
		\caption{Color Jitter}
	\end{subfigure}
	\hfill
	\begin{subfigure}[b]{0.16\textwidth}
		\centering
		\includegraphics[width=\textwidth]{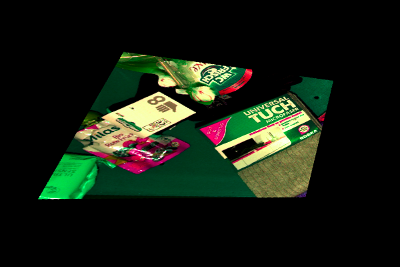}
		\caption{Combined}
		\label{fig:combined_aug}
	\end{subfigure}
	\caption{Visualization of the augmentations utilized for synthetic view training. \Cref{fig:combined_aug} shows the combination of all augmentations as used in practice for training a synthetic view descriptor model.
	While each augmentation is guaranteed to be applied, the individual parameters, e.g., scale of distortion, angle of rotation, crop size and location, etc. are still randomly selected each iteration.}
	\label{fig:three graphs}
\end{figure}

We carefully select augmentations, which reflect the desired invariance properties stated above, as follows: \textit{affine transformation} induce rotations and scale changes (zoom out), \textit{perspective distortions} mimic view changes, \textit{resize\&crop} implies scale changes (zoom in), and lastly, \textit{color jitter} affects brightness, contrast, hue, and saturation of the image, hence the lighting conditions of the scene.
We show an illustration of these augmentations in Fig.~\ref{fig:three graphs}.
We utilize \textit{torchvision library} \cite{Pytorch2019} of Pytorch as reference implementations.

\citet{Adrian2022} already demonstrated the improved performance based on image augmentations for the training of dense visual descriptors on datasets utilizing geometric correspondences.
In our work, each augmentation is not only helpful, but relevant to the ability of the model to successfully learn an invariant descriptor space.
See Sec.~\ref{sec:augmentations_abl} for an ablation on the respective impact of each augmentation on the overall performance.

\subsection{Loss Function}
\label{sec:loss}
Following \citet{Chen2020} we adopt the NT-Xent loss, for learning the dense descriptor representations.
For a pair of corresponding descriptors $\{d_i, d_j\}$, obtained from images $I'$ and $I''$ in a minibatch of size $M$, we compare their distance to the distance of $d_i$ to all other sampled descriptors in the given minibatch arriving at the following individual loss term:
\begin{equation}
l_{i,j} = -\log \frac{\exp(\dist{d_i, d_j}/\tau)}{\sum_{k=1; k\neq i}^{2M}  \exp(\dist{d_i, d_k}/\tau)}.
\label{eq:ntxent}
\end{equation}
with temperature parameter $\tau$ which we fix to $\tau=0.07$ throughout this paper following \citet{Adrian2022}.
We choose the metric $\dist{\cdot}$ to be the cosine similarity between descriptors $d_i$ and $d_j$.
As the cosine similarity expects normalized vectors we normalize the descriptors $d$.
The total loss is given as the mean over all individual loss terms, cf.~\citet{Chen2020}.

Note that this loss, together with our correspondence sampling, does not distinguish background from objects, nor does it explicitly address multiple object classes and instances, as opposed to the approach by \citet{Florence2018}.
Instead, we follow the method of \citet{Adrian2022} and sample correspondences uniformly in image plane and assume that every pixel is unique.
This method is tailored to datasets depicting densely packed scenes with single object instances, for example, a heap of objects in a bin-picking scenario.
The learned descriptor space does not imply semantic information on object classes or background, but still provides consistent keypoint detection and robust tracking performance, which is essential for downstream tasks.

\section{Comparison of Training with Synthetic and Geometric Correspondence}
\label{sec:results}
In this section we show an in-depth comparison between training with geometric and synthetic correspondence.
We also investigate the invariance of descriptors obtained from a network trained with synthetic correspondence with respect to object-camera relative transformations.

In all our evaluations we utilize a pretrained ResNet-34 with 8-stride output as used by \citet{Florence2018}, which yields an upsampled output matching the resolution of the input.
In the supplementary material we give a brief introduction into the baseline training method we use in our comparisons relying on  geometric correspondence by \citet{Adrian2022}.

We recorded a dataset consisting of a set of registered RGBD sequences, to enable comparison between both approaches.
Despite the availability for registered image pairs, the synthetic view training only uses single RGB images for both training and validation.
However, the camera poses help with the evaluation as they allow us to generate ground truth pixel matches across any two images of the same static scene without the need for manual labeling.
The dataset consists of eight scenes with various object configurations, with every scene containing only one instance per object, and every object is visible to some extent in every frame.
The scenes are recorded with a robot wrist mounted camera while the robot arm follows a predefined trajectory keeping the objects in view.

Both approaches are evaluated on the same ground-truth image pairs and correspondences.
For robustness, we perform a k-fold cross-validation, that is, each scene from our total dataset was once used as test set.
One scene is chosen as validation set, with the remaining 6 scenes used for training.
The averaged results are reported.
We use the same loss function, training parameters and augmentations for both approaches, with the exception that augmentations are chosen with $50\%$ probability for the geometric training, as it yields better results.
For synthetic training, each augmentation is always used.
Training details and an ablation study of using different augmentation probabilities is given in the supplementary materials.

\subsection{Keypoint Tracking Performance}
\label{subsec:keypoint_tracking}

Given a common dataset, measuring keypoint tracking accuracy provides a task-agnostic comparison between both training methods.
We define a keypoint as a location in image plane $k_i = (u, v)$.
Each keypoint is associated with a unique descriptor $d\in \mathbb{R}^D$ and the keypoint tracking problem is defined as finding the pixel location $k_i^*$ closest to $d$ in the descriptor image $I_d$ such that $k_i^* = \arg \min_{k_i} \mbox{dist}(I_d(k_i), d)$. 

\textbf{Evaluation.} 
From the test set we sample $1000$ image pairs $\{A, B\}$ representing alternative views of the same scene.
For each image pair we sample $200$ keypoints $\{k_i^A, d_A\}$, where $~d_A=f(A;\theta)(k_i^A)$ is located on one of the objects in image $A$.
We also recorded keypoint locations in image $B$, $\{k_i^B\}$, such that every pair $\{k_i^A, k_i^B\}$ projects from pixel to world coordinates.
Then, we solve the keypoint tracking problem for every descriptor $d_A$ in image $B$ and record the pixel error $e = \lVert k^B_i - k_i^* \rVert_2$.
The network parameters $\theta$ are either obtained by geometric \cite{Adrian2022} or synthetic correspondence training proposed in Sec.~\ref{subsec:single_view_training}.

\begin{figure}[h]
	\centering
	\begin{subfigure}{0.48\textwidth}
		\includegraphics[width=1.0\textwidth]{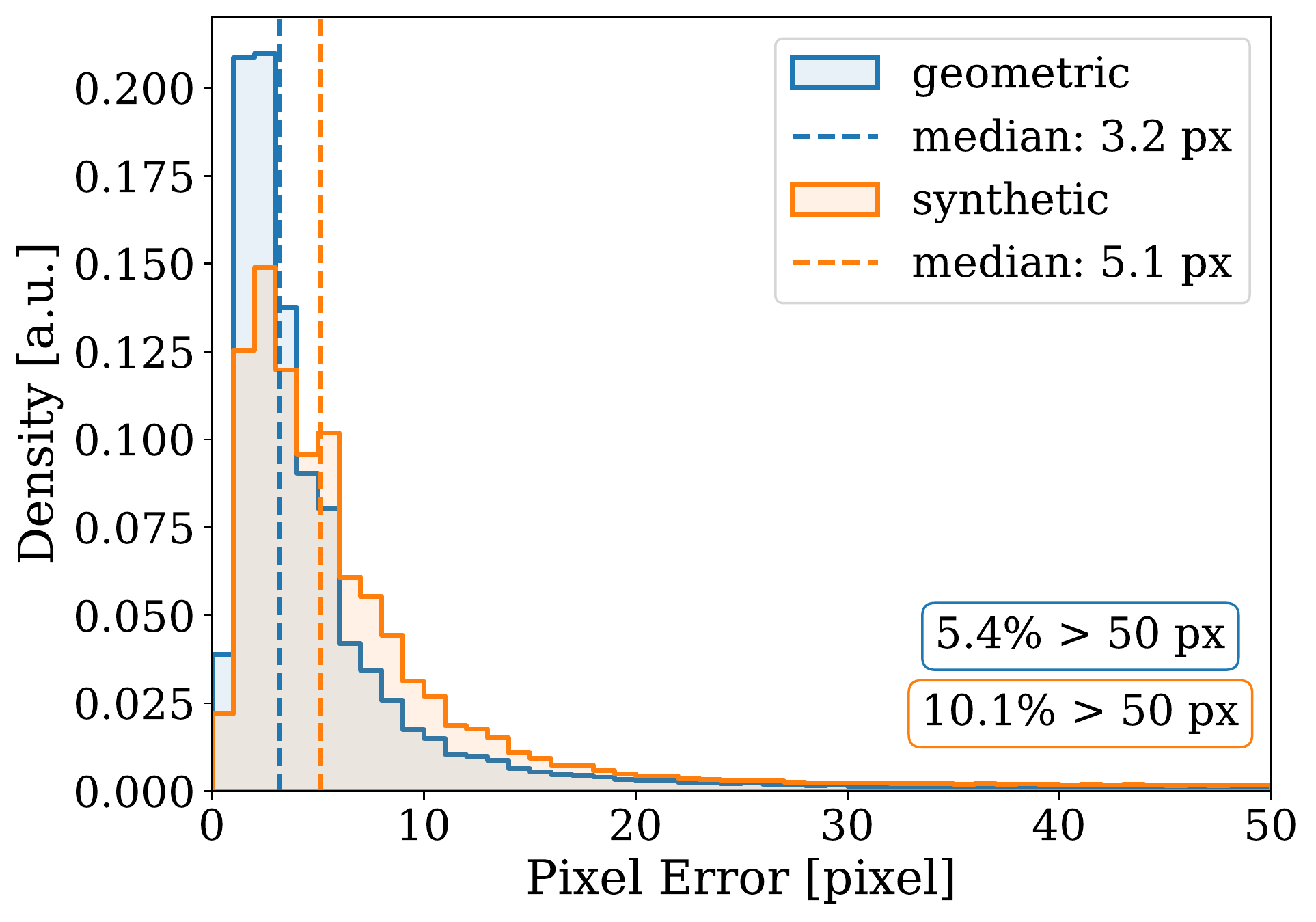}
		\caption{}
		\label{fig:pixel_errors_baseline}
	\end{subfigure}
	\hfill
	\begin{subfigure}{0.45\textwidth}
		\includegraphics[width=1.0\textwidth]{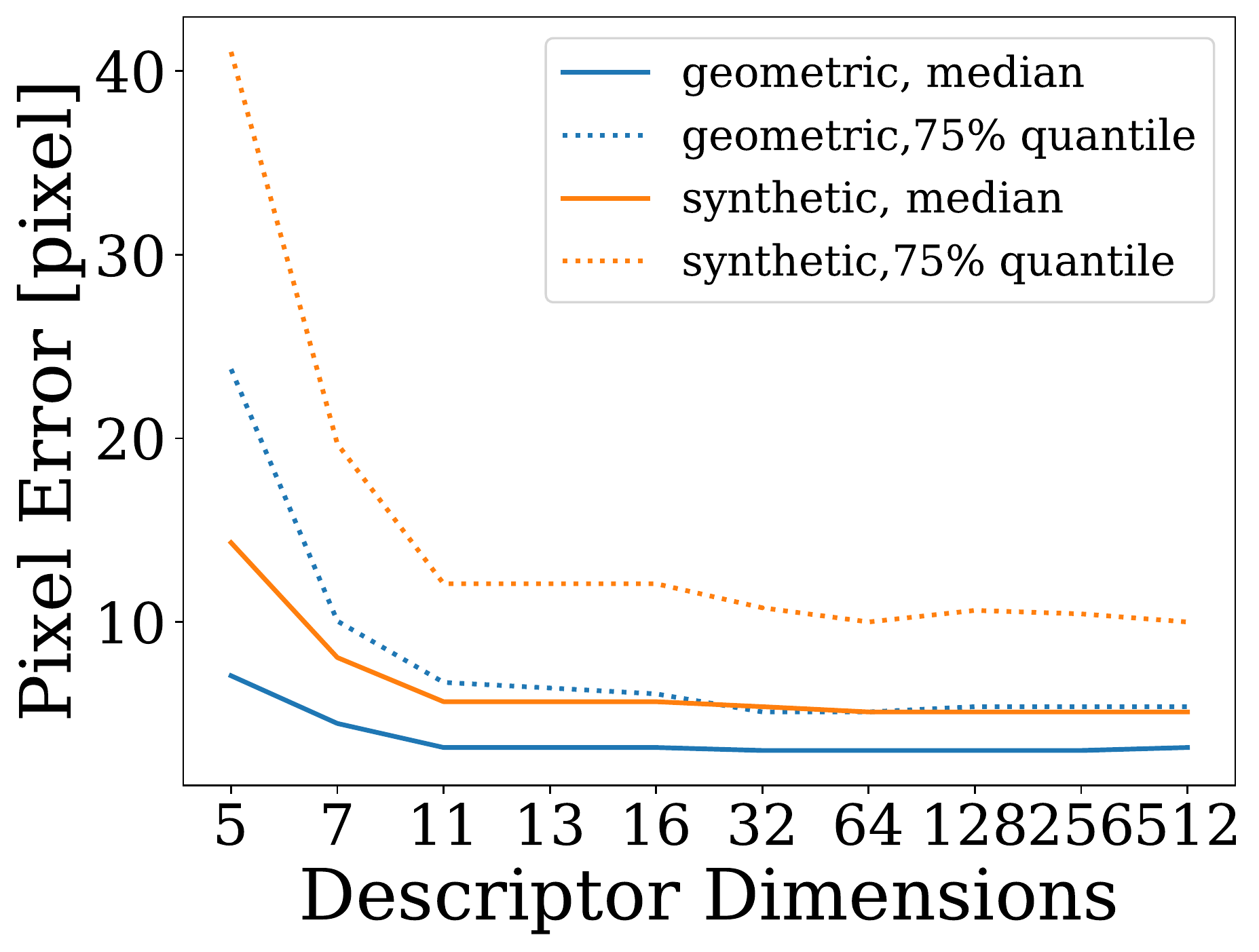}
		\caption{}
		\label{fig:pixel_errors_dim_study}
	\end{subfigure}
	
	\caption{(a) Pixel error distributions obtained using synthetic (orange) and geometric (blue) correspondence training.
		(b) Median and 75\% quantile of the pixel error distribution for both training approaches and different descriptor dimensions.}
\end{figure}

\textbf{Results.}
Fig.~\ref{fig:pixel_errors_baseline} shows the distribution of pixel errors for the two training approaches.
The median of the corresponding distribution is highlighted as a dashed line. 
With a difference of only $1.9$px in median pixel error, the synthetic correspondence performs competitively to the geometric training.
The percentage of pixel errors that are larger than 50 pixels are $10.1\%$ and $5.4\%$ respectively, as indicated in the bottom right part of Fig.~\ref{fig:pixel_errors_baseline}.

In Fig.~\ref{fig:pixel_errors_dim_study} we compare the median and the 75\% quantile of the two training methods with respect to the descriptor dimensions $D$.
The results were obtained without k-fold cross-validation.
For both training approaches the median pixel error decreases for larger descriptor dimensions.
For dimensions larger than $9$ the median pixel error decreases only marginally.
In contrast, the 75\% quantile error still increases further until $D=64$, before improvements saturate.
For additional insights into these results we refer to the supplementary material.

\subsection{Invariance Tests} 
\label{sec:invariance_test}
\begin{figure}[h] 
	\centering
	\begin{subfigure}{0.32\textwidth}
		\includegraphics[width=1.0\textwidth]{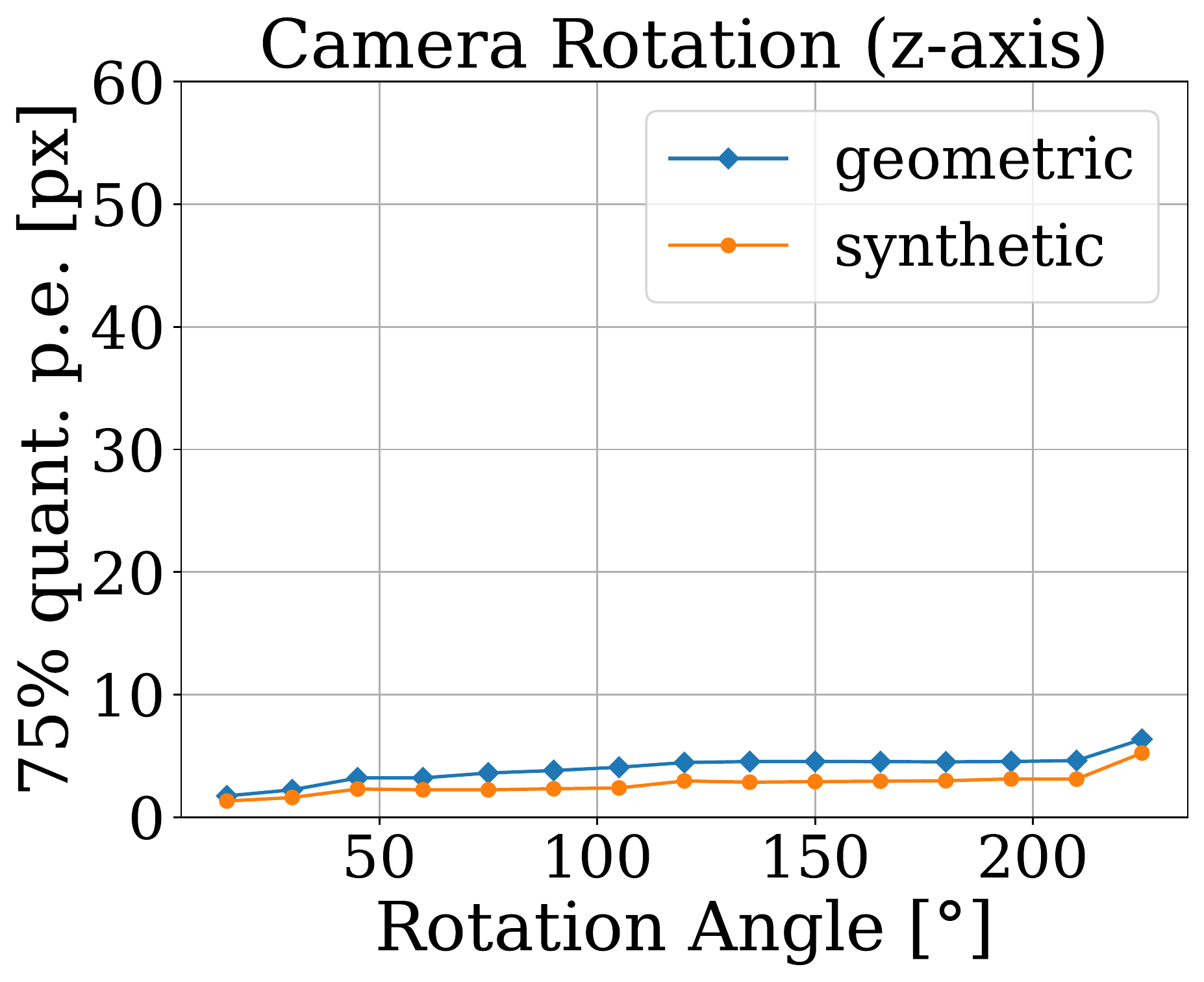}
		\caption{}
		\label{fig:pe_rot_z}
	\end{subfigure}
	\hfill
	\begin{subfigure}{0.32\textwidth}
		\includegraphics[width=1.0\textwidth]{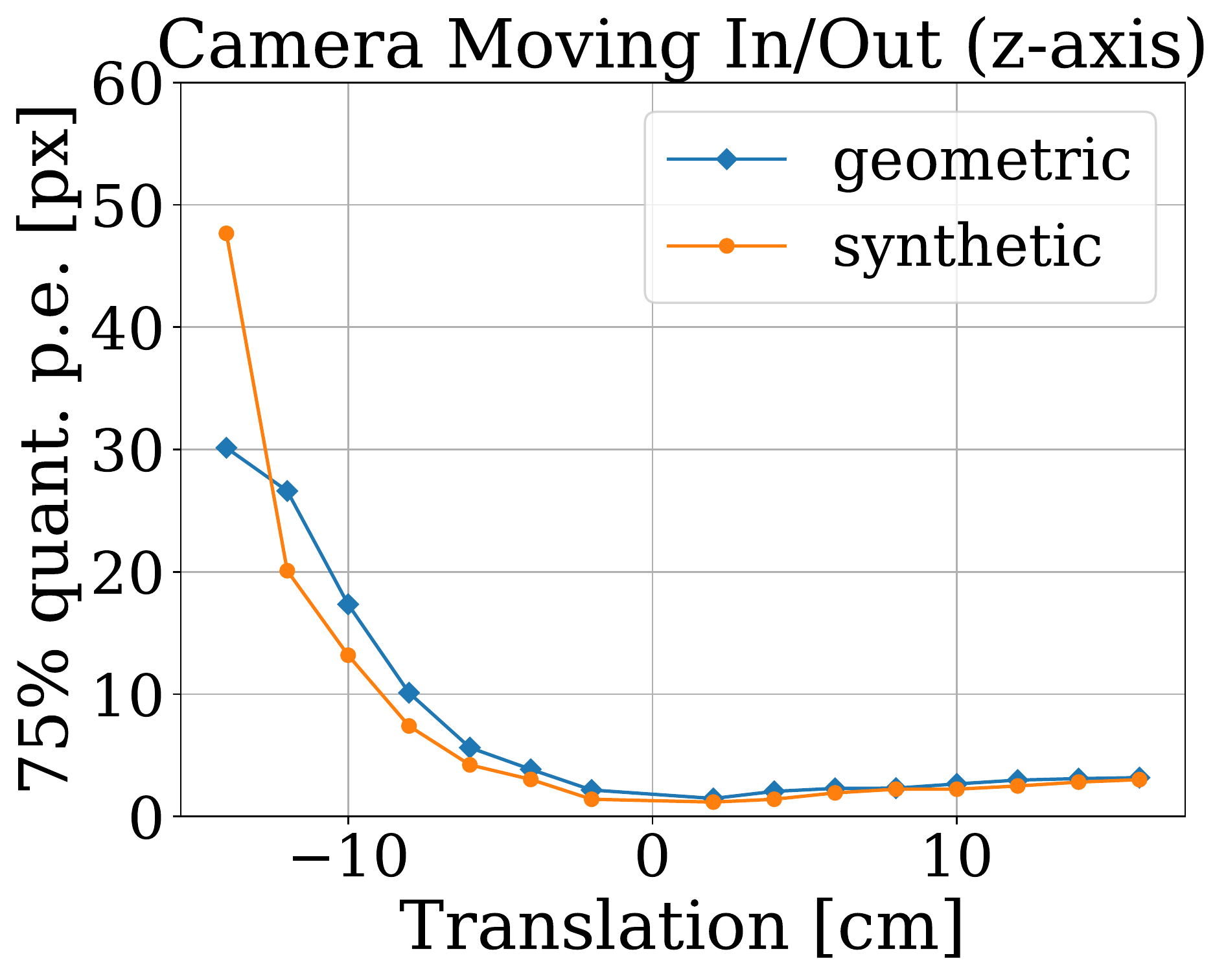}
		\caption{}
		\label{fig:pe_trans_z}
	\end{subfigure}
	\hfill
	\begin{subfigure}{0.32\textwidth}
		\includegraphics[width=1.0\textwidth]{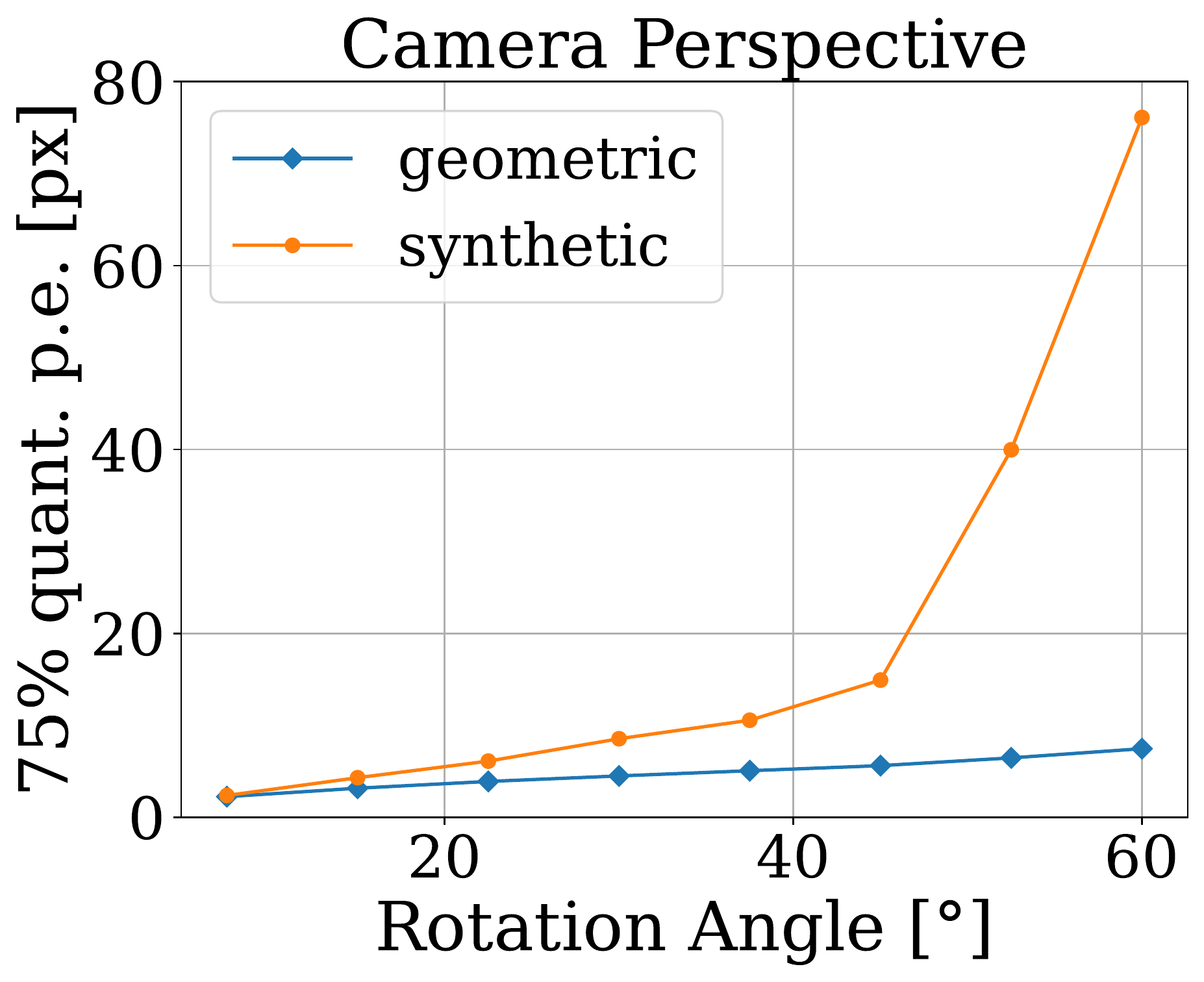}
		\caption{}
		\label{fig:pe_tilt_x}
	\end{subfigure}
	\caption{75\% quantile pixel error between two different images for the geometric (blue) and synthetic (orange) correspondence training approach
		for three different isolated camera movements between those images in varying magnitude: 
		(a) camera rotates around z-axis,
		(b) camera moving closer and further away from the objects,
		(c) camera moving on a sphere in x-direction around the objects, facing the objects.
		Note the scale of the y-axes.}
	\label{fig:invariance_tests}
\end{figure}

In the following, we wish to answer the question: how well does training with image augmentations proposed in Sec.~\ref{sec:augmentations} generalize to physical camera transformations?
For this purpose, we recorded different test scenes with a wrist-mounted camera and the following specific camera movements: (i) changing camera perspective (camera tilting), (ii) translation along the camera z-axis (zooming in and out), and (iii) camera rotation along the camera z-axis (see Sec.~\ref{app:invariance_camera_movements} for details).
We compare the performance of a network trained with synthetic and one with geometric correspondence both trained on the same data as described in Sec.~\ref{subsec:keypoint_tracking} with 64 descriptor dimensions and affine, perspective and resize\&crop augmentations.
As in the previous section we compute the 75\% quantile pixel error and use $1000$ keypoints per image pair, fix the base image $A$ and only vary image $B$, which shows the changing camera views.
The results for three different types of transformations are compiled in Fig.~\ref{fig:invariance_tests}.

It can be seen that training with synthetic correspondence generalizes well to a large range of camera transformations, especially to those parallel to the object plane (Fig.~\ref{fig:pe_rot_z}, Fig.~\ref{fig:pe_trans_z}).
These physical camera transformations are very similar to the affine and resize\&crop augmentations used during training.
For generalizing to the perspective transformations with angles above $45^\circ$ degrees, as shown in Fig.~\ref{fig:pe_tilt_x}, the synthetic correspondence training shows a clear deterioration in performance.
As the perspective changes, occluded parts become visible and vice versa. 
This physical transformation is not well captured by the synthetic augmentations for larger angles.

\subsection{Augmentations}
\label{sec:augmentations_abl}
\begin{figure}[h] 
	\centering
	\begin{subfigure}{0.32\textwidth}
		\includegraphics[width=1.0\textwidth]{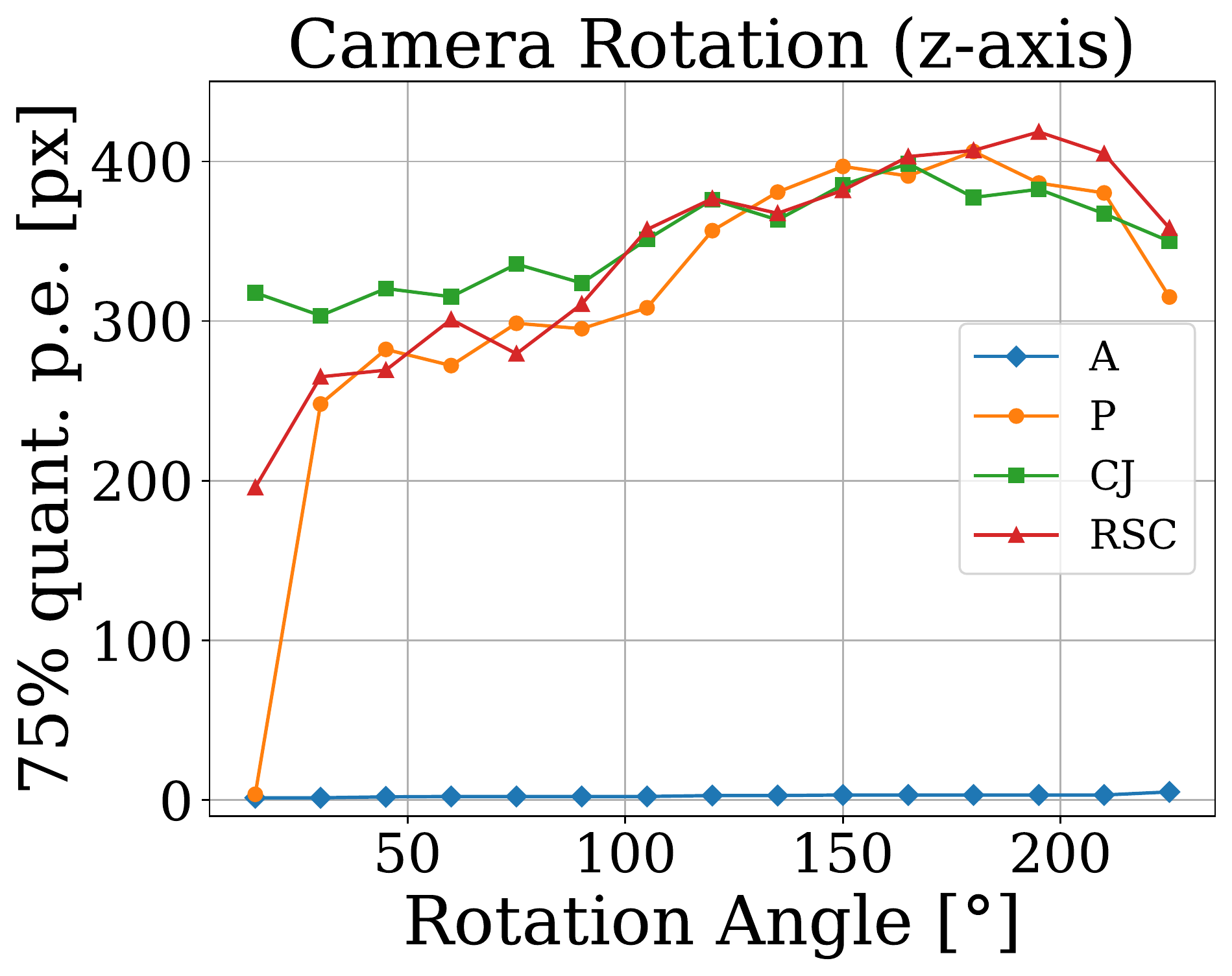}
		\caption{}
		\label{fig:abl_rotation_base_augs}
	\end{subfigure}
	\hfill
	\begin{subfigure}{0.32\textwidth}
	\includegraphics[width=1.0\textwidth]{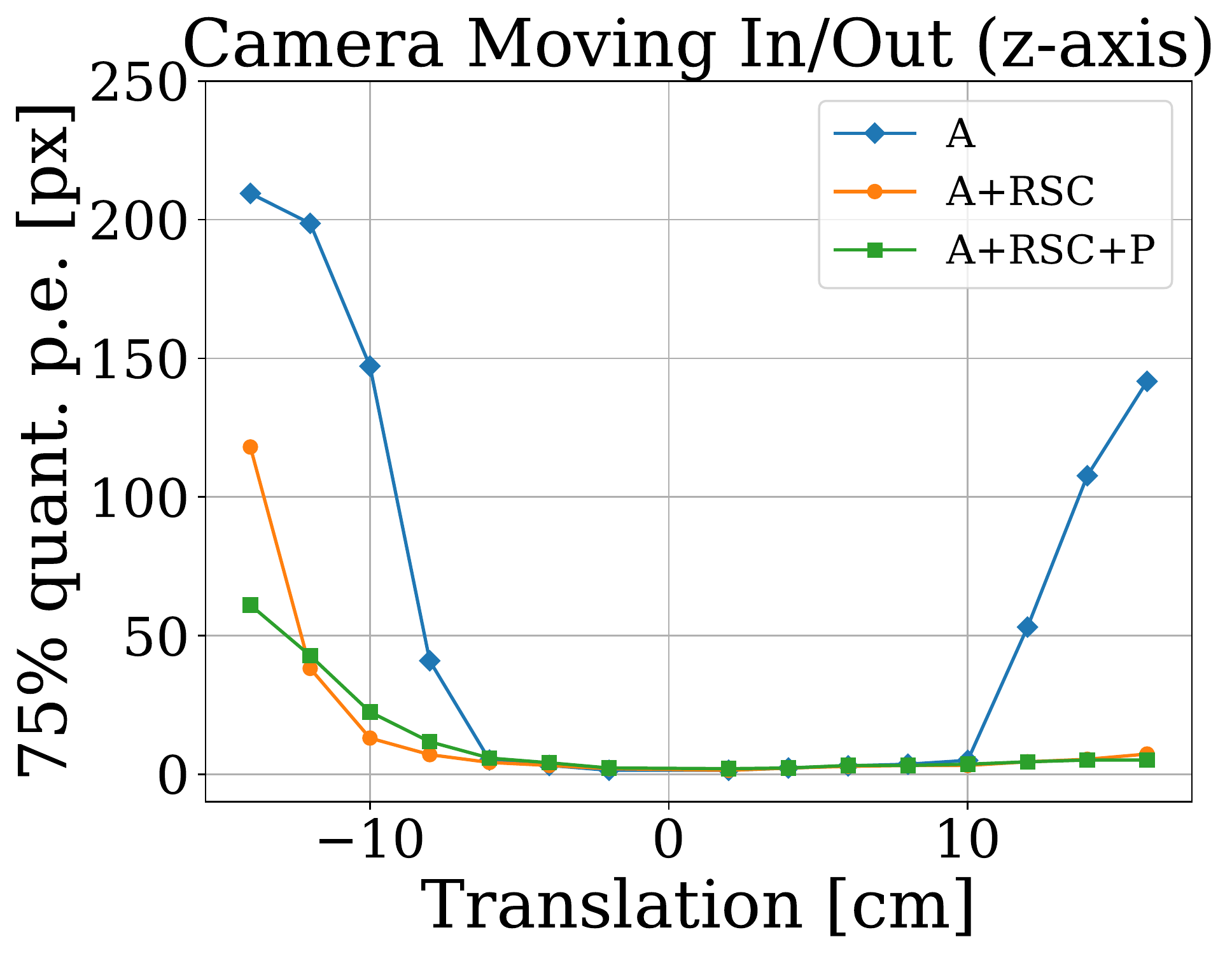}
	\caption{}
	\label{fig:abl_zoom_out_incremental}
	\end{subfigure}
	\hfill
	\begin{subfigure}{0.32\textwidth}
		\includegraphics[width=1.0\textwidth]{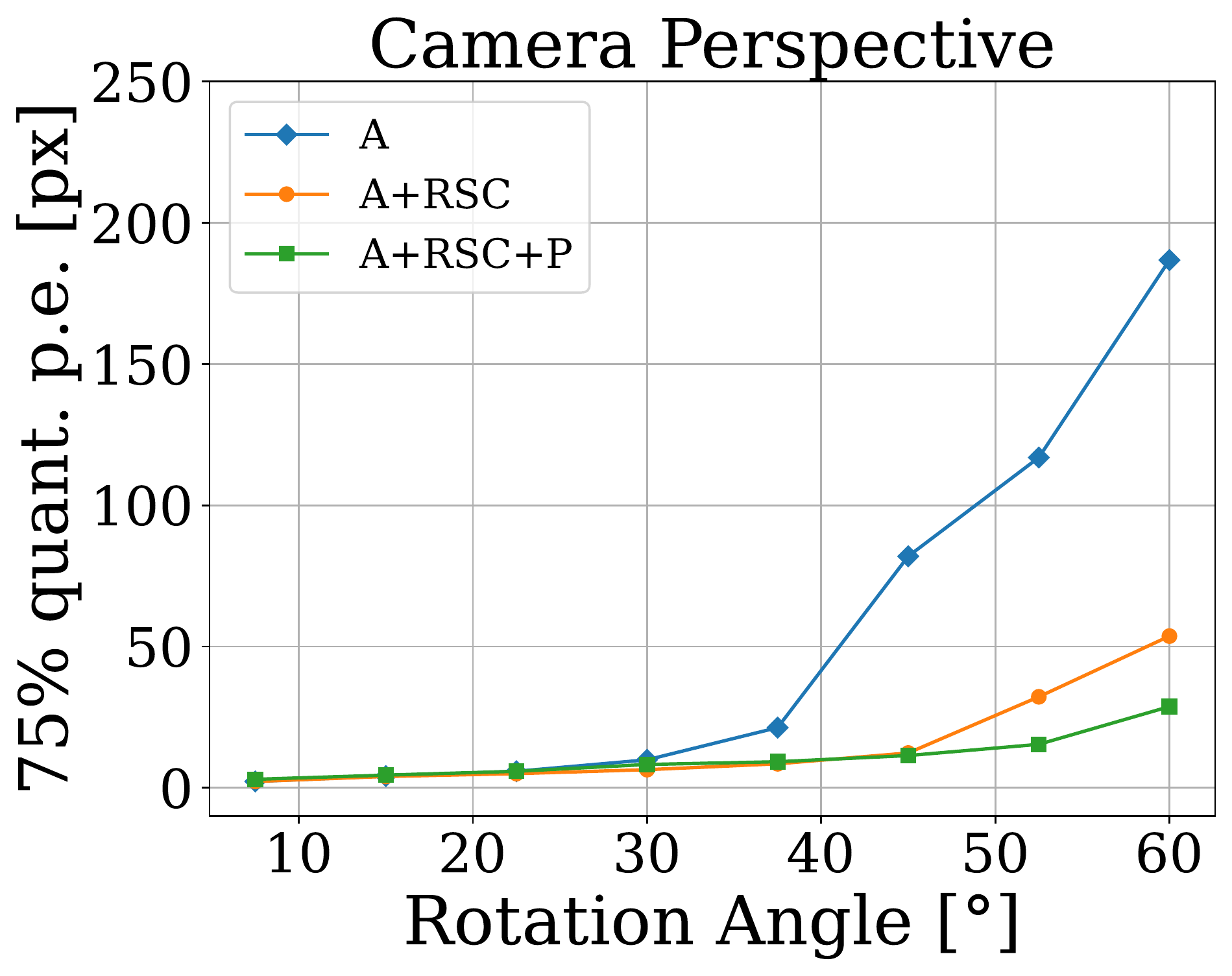}
		\caption{}
		\label{fig:abl_xzy}
	\end{subfigure}
	\caption{75\% quantile pixel error for different combinations of augmentation. \textbf{A}ffine, \textbf{P}erspective, \textbf{C}olor \textbf{J}itter, \textbf{R}e\textbf{s}ize \& \textbf{C}rop, with respect to different tasks: 
		(a) camera rotates around z-axis,
		(b) camera moving closer and further away from the objects,
		(c) camera moving on a sphere in x-direction around the objects, facing the objects.
		Note the scale of the y-axes.}
	\label{fig:ablation_invariance_tests}
\end{figure}

Complementing the findings in \cref{sec:invariance_test} we study the influence of different augmentations for the synthetic correspondence  training on the ability of the network to generalize to different camera transformations.
Fig.~\ref{fig:ablation_invariance_tests} shows the 75\% quantile of the pixel error distribution for the synthetic correspondence training with different augmentations.
We find that \textit{affine transformations} are most critical, as only a model trained with it shows invariance to rotations, see Fig.~\ref{fig:abl_rotation_base_augs}.
This matches the expectations as standard CNN are by default not invariant to rotations.
Nevertheless, we find that both \textit{resize\&crop} and \textit{perspective distortion} both further improve the performance of just affine transformations.
In particular, for camera movements that induce perspective distortions and scale changes, see Fig.~\ref{fig:abl_xzy} and Fig.~\ref{fig:abl_zoom_out_incremental}, the error decreases considerably.
Lastly, we find that \textit{color jitter} further reduces the overall mean pixel error from $19.4$ to $17.1$ pixel.
The improvement appears modest, but we note that our test dataset was recorded at the same time as train and validation, and the lighting conditions of the scene were not explicitly altered.
For a complete table of all the combinations of augmentations, see the supplementary material.

\section{Grasping Experiment with Fix-mounted Camera}
\label{sec:experiment}

We demonstrate a robotic bin-picking experiment that relies on dense visual descriptors for defining grasp preferences.
We use a 7-DoF Franka Emika Panda arm with a suction gripper mounted on the end-effector, see Fig.~\ref{fig:bin_picking_setup}.
Our setup uses a fix-mounted Zivid One+ camera above the bin in a robotic cell.
Training a descriptor network using this setup prevents the use of geometric correspondences.
Instead, we show that our proposed method can be trained on a setup as it is often present in real world applications and prove that the keypoint tracking accuracy is good enough for guiding a generic grasping method by human annotated grasp preferences.

In the experiment, we consider picking ten different types of objects from one bin.
However, using a suction gripper the objects are difficult to grasp at certain locations: some have cutouts on the packaging, transparent or foldable parts, and uneven surfaces (see supplementary material).
In our experiments, a purely model-free grasp pose generator often predicts poses at these challenging parts of objects, ultimately reducing picking performance.

\begin{figure}[h]
	\centering
	\begin{subfigure}{.24\textwidth}
		\includegraphics[width=\textwidth]{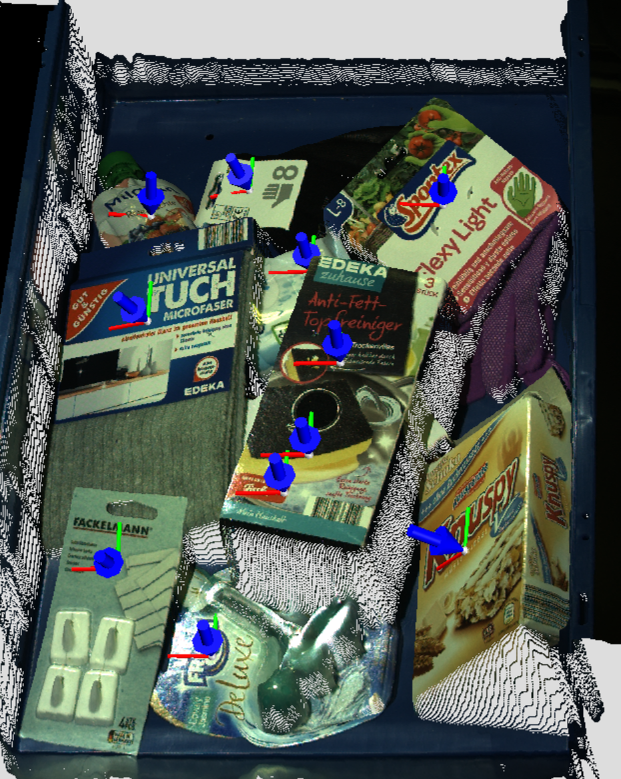}
		\caption{}
		\label{fig:grasp_illustration_rgb}
	\end{subfigure}
	\hfill
	\begin{subfigure}{.24\textwidth}
		\includegraphics[width=\textwidth]{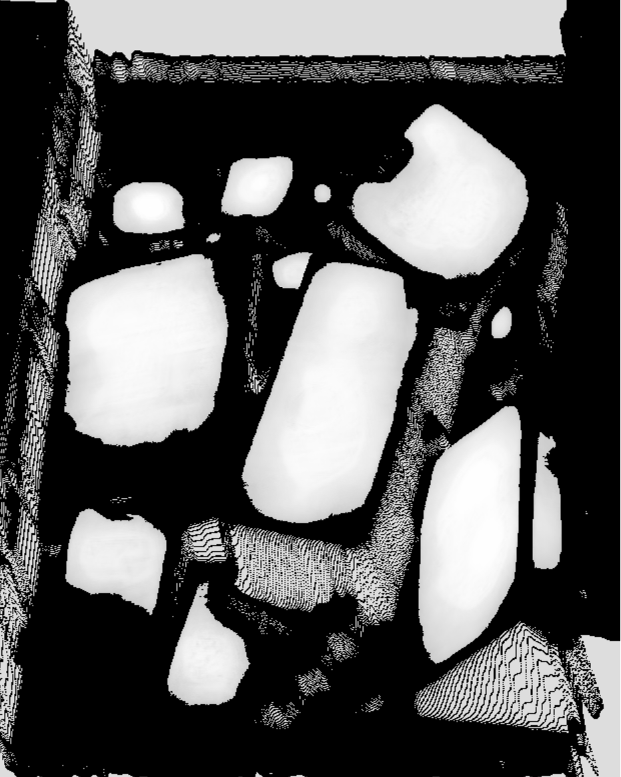}
		\caption{}
		\label{fig:grasp_illustration_quality}
	\end{subfigure}
	\hfill
	\begin{subfigure}{.24\textwidth}
		\includegraphics[width=\textwidth]{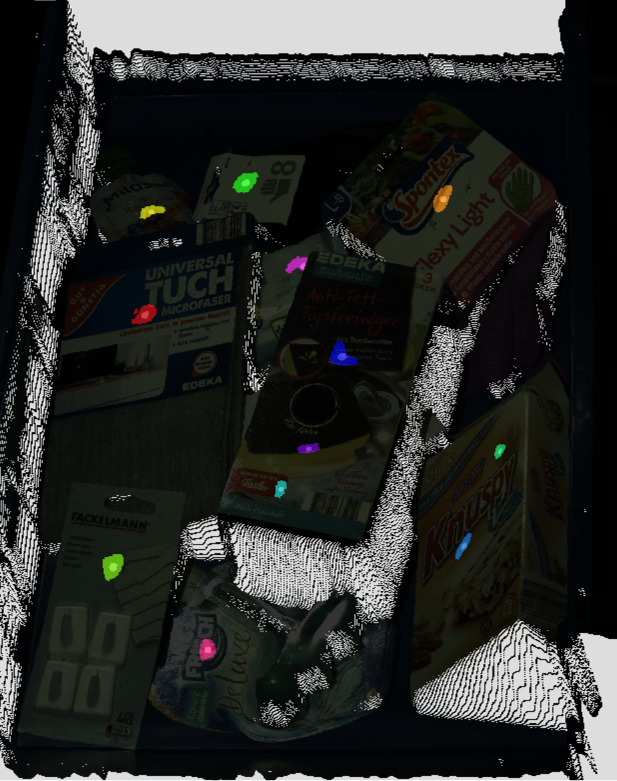}
		\caption{}
		\label{fig:grasp_illustration_heatmaps}
	\end{subfigure}
	\begin{subfigure}{.24\textwidth}
		\includegraphics[width=\textwidth]{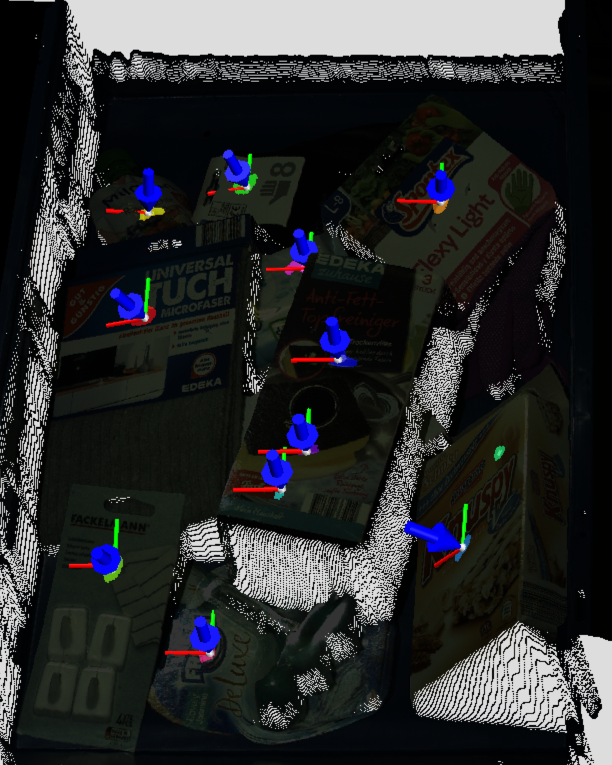}
		\caption{}
		\label{fig:grasp_illustration_heatmaps_grasps}
	\end{subfigure}
	\caption{Illustration of the grasping experiment. (a) The RGB images with the predicted grasp poses. (b) The graspable areas generated using RGB and depth images. (c) The predicted keypoint heatmaps on graspable areas. (d) The predicted grasp poses overlayed on the heatmaps.}
	\label{fig:grasp_illustration}
\end{figure}

We hypothesize that introducing human domain knowledge to aid grasp pose selection will lead to higher grasp success rates.
Human domain knowledge is considered by highlighting parts of the RGB image where grasp poses are preferred.
To do so, we show a small number of RGB images depicting the objects in different configurations in the bin and ask the human to click on pixel locations corresponding to a preferred grasp location.
We track these descriptors to generate a \emph{preference heatmap}, as shown in Fig.~\ref{fig:grasp_illustration_heatmaps}.
The heatmap, in contrast to discrete single-pixel correspondences, enables intuitive consideration of matching uncertainty in the form of distance in descriptor space.
Furthermore, it provides a more flexible and quantifiable basis for combination with grasp detectors.

To generate the final set of grasp pose candidates we intersect the grasp preference heatmap with the detected graspable areas identified by the model-free grasp detector (Fig.~\ref{fig:grasp_illustration_quality}).
While we make no specific assumption regarding which grasp detection method to use, we employ in the shown experiment a dense pixel-wise graspability estimation based on a fully convolutional neural network with RGB-D input, specifically UNet~\cite{ronneberger2015u} trained on annotated pixel-wise labels of expected graspability for a wide range of different bin picking scenes.
For the example image, the resulting poses are shown in Fig.~\ref{fig:grasp_illustration_rgb} and Fig.~\ref{fig:grasp_illustration_heatmaps_grasps}.

Note that the descriptor network is trained purely from the set of RGB images of the bin, including the objects in random configurations, recorded with the overhead camera.
We require a single instance of the objects to be present in the bin.
For more details on the experiment setup and heatmap generation we refer to the supplementary material.

\subsection{Quantitative Evaluation}

We evaluate the benefit of the proposed method for robotic bin picking based on a set of sixteen manually annotated scenes.
The scene images include the ten known objects, but in different configurations compared to the training set.
For each evaluation image, we manually annotated object instances, pixel-wise graspable areas, and those areas that correspond to selected descriptors.
In total, the evaluation dataset contains 131 graspable objects of which 103 have visible descriptor spots.

We compare our descriptor-based grasping approach with a purely model-free approach that directly uses the graspable areas without accounting for descriptors.
With this study, we investigate two questions regarding the descriptor-based approach.
First, how effective is the approach to re-identify descriptor spots compared to grasping at these spots by chance, considering that the descriptors indicate the best way of grasping the respective object?
And second, is there a considerable negative effect on the amount of objects that can be grasped, e.g., due to missing to identify graspable areas?

Table~\ref{tab:evaluation} shows a summary of the results.
\emph{Success Rate} denotes the number of successful grasps compared to all grasps attempted.
\emph{Descriptor Success} denotes the number of grasps at those spots marked as desired grasping points compared to all grasps attempted.
\emph{Object Hits} denotes the percentage of all graspable objects for which at least one feasible grasp pose has been found irrespective of descriptors, including objects without visible descriptor spots.
\emph{Descriptor Hits} denotes the percentage of objects for which a grasping pose has been found at the respective descriptor spot, only including objects with visible descriptors.
We consider descriptor spots with a tolerance of around 1cm, which corresponds to the radius of the suction gripper.

\begin{table}[h]
	\caption{Evaluation results on individual, annotated scene images similar to Fig.~\ref{fig:grasp_illustration}.}
	\label{tab:evaluation}
	\begin{tabular}{lrrrr}
		                  & Success Rate & Descriptor Success & Object Hits & Descriptor Hits \\ \hline
		Preference (ours) &       98.9\% &             91.1\% &      63.4\% &          78.6\% \\
		Baseline          &       79.9\% &             50.4\% &      78.6\% &          68.0\%
	\end{tabular}
\end{table}

It can be concluded from Table~\ref{tab:evaluation} that using the proposed method to encode grasp preferences is effective as it significantly increases the amount of grasps at desired spots from $50.4\%$ to $91.1\%$.
Due to the challenging object geometries, this helps to raise the overall grasp success rate from $79.9\%$ to $98.9\%$.
The expected downside, however, is that the proposed method finds grasp poses only for a smaller amount of objects, $63.4\%$ instead of $78.6\%$.
Still, this includes objects that have no desired grasping spot visible.
In some applications it can be the desired behavior to not propose a grasp pose for objects if they cannot be grasped at the preferred location.
When only considering which preferred grasping spots have been covered by grasp poses, our method manages to find grasps for $78.6\%$ instead of $68.0\%$ of the visible spots.
In consequence, we conclude that there is only a moderate negative effect due to limiting grasping to descriptor spots which can indeed be beneficial for some applications.

\section{Limitations}
In the following we share more insights on the limitations of our method and discuss future research. \\
\textbf{Variance in camera poses.}
As seen in Fig.~\ref{fig:invariance_tests}, the synthetic training ensures that descriptors are stable within a limited margin of camera transformations relative to the object.
For example, in Fig.~\ref{fig:pe_tilt_x}, for viewing angles steeper than $45^\circ$, accuracy deteriorates.
This imposes a limit on descriptor consistency across images with large differences in object poses.\\
\textbf{Changing Environment.}
A change in environment (background, lighting) between training and inference time may have a negative influence on keypoint tracking performance.
It is expected that augmentations such as color jitter and, if masks are available, background randomization can mitigate these effects.
We will investigate these aspects in future research.\\
\textbf{Generalization to unseen objects.}
Although \citet{Florence2018} demonstrated capabilities to generalize intra-class instances, our work focuses on instance specific descriptors.
Given our outlined training setup, it works best on known objects from the training set.\\
\textbf{Object Edges.}
We observe worse descriptor consistency across images for keypoints located at the edges of objects or close to parts that are occluded by other objects.
In setups where object masks are present background randomization could reduce this effect.

\section{Conclusion}
In this paper we proposed a novel training method for learning dense visual descriptors based on image augmentations for robotic manipulation.
The evaluation shows that overall our proposed method is competitive to the existing geometric training approach.
For physical transformations like changing the camera perspective on the scene, which are harder to mimic by augmentations, the training method using geometric correspondences shows superior performance.
Being aware of these limitations our proposed method is expected to perform well for setups where objects are mostly altered by translations, rotations parallel to the camera plane, or are slightly tilted.
As this is often the case for random heaps of objects in a bin, our method is especially suitable for such setups that are constrained by a fix-mounted camera.
Finally, we demonstrated the use of our method in a realistic grasping experiment to increase grasp success rates by human annotated grasp preferences.

\clearpage
\bibliography{example}  

\clearpage
\appendix
\section{Overview}
In the following, we provide additional technical details and supporting results.
The sections are organized as follows:
\begin{itemize}
	\item Sec. \ref{sec:training_with_geometric_correspondences} reviews how training with geometric correspondences is conducted.
	\item Sec. \ref{sec:apx_training_details} provides further details for training models.
	\item Sec. \ref{sec:apx_datasets} offers an overview and example images of our datasets.
	\item Sec. \ref{sec:apx_ablation_augmentation} discusses further results from the augmentation ablation study,
	\item Sec. \ref{sec:apx_grasping} details grasp preference heatmap generation and the objects used in the grasping experiment.
\end{itemize}

\section{Training with Geometric Correspondences}
\label{sec:training_with_geometric_correspondences}
While originally introduced by \citet{Florence2018}, we utilize the adapted method by \cite{Adrian2022} for training without masks in multi-object settings.

The training relies on sampling a set of corresponding pixels in image $A$ and $B$, where both images observe the same static scene and objects, but from different view points.
By employing a contrastive loss, the descriptors for each pixel pair are trained to have the same embedding, while separating all other pixels in latent space.
The view-invariance of the descriptors is the consequence of utilizing images with different view points.

Geometric correspondence training exploits the geometric prior provided by a registered RGBD sequence.
As the relative pose between any two images in the sequence is known, and given the depth and camera information, one can establish the per-pixel correspondence between each image pair, allowing for straight-forward sampling correspondences.

In practice, depth data can be noisy or incomplete.
For example, structured light cameras struggle with transparent or black surfaces, and with higher measurement uncertainty around edges of objects.
Thus, \citet{Florence2018} perform a 3D-reconstruction of the scene, to render synthetic depth images which are complete and denoised, albeit not perfect ground-truth.
In the original approach, which focuses on training with singulated objects, an automatic or manual mask generation of the object is performed.
As we deal with multi-object scenes, we follow \cite{Adrian2022} and instead record scenes with multiple objects present and do not compute any masks.
Hence, we sample correspondences anywhere in the image and do not differentiate between object or background.

Instead, to sample correspondences we first prune the correspondence map from image $A$ to $B$ by occlusion and field of view masking, then we sample a set of $N$ pixel correspondences.
As we apply augmentations, the process can shift or remove pixels from either image. We need to account for this in the sampling process.
Given a set of sampled correspondences, we employ the same loss as described for our synthetic view training.

\section{Training Details}
\label{sec:apx_training_details}
The training settings, see Table \ref{tab:general_settings}, are shared for both the geometric and synthetic training, with exceptions specified below.
They follow the findings made by \cite{Adrian2022} for geometric training.
They are used for all experiments shown, unless specified otherwise.

\begin{table}[ht]
\begin{center}
\caption{Default training settings for both geometric and synthetic view training.}
\label{tab:general_settings}
\begin{tabular}{lc}
	\toprule
	Parameter & Setting \\
	\midrule
	Latent dimension (default) & 64 \\
	Temperature (NTXent) & 0.07 \\
	Optimizer & Adam \cite{Kingma14} \\
	Learning rate & 0.0003 \\
	Number of correspondences per image pair & 2048\\
	Batch size & 2 \\
	Batches per epoch & 500 \\
	Validation every n epochs & 1\\
	Total epochs & 250 \\
	\bottomrule
\end{tabular}
\end{center}
\end{table}

We perform validation after each epoch and retain the checkpoint with the best score.
The score is evaluated with respect to the area under the curve (AUC) of the PCK@K (percentage of correct keypoints).
PCK@K is determined by taking a set of predictions and calculating the pixel error $e$ with respect to the corresponding ground-truth pixels.
The percentage is given by the number of predictions with a pixel error $e < K$.
We evaluate the AUC for the range $K\in[1\cdots100]$.

Both approaches, geometric and synthetic, are trained with the same augmentation parameters as listed in Table~\ref{tab:augmentation_settings}.
The major difference is that for synthetic training, we sample each augmentation with probability ${p=1.0}$ , whereas for geometric training each augmentation is sampled with ${p=0.5}$.
We found that the latter performed better.

\begin{table}[ht]
	\begin{center}
		\caption{Settings for augmentations with respect to the Torchvision library implementation.}
		\label{tab:augmentation_settings}
		\begin{tabular}{lcc}
			\toprule
  			\multicolumn{2}{c}{Parameter} & Setting \\
			\midrule
  			\textbf{Color Jitter} & & \\
  			& Brightness & $0.2$ \\
  			& Contrast & $0.2$ \\
  			& Saturation & $0.2$ \\
  			& Hue & $0.2$ \\
  			\textbf{Affine} & & \\
			& Rotation Angle & $[0\dots359]$ \\
			& Scale & $[0.5\dots1.0]$ \\
  			\textbf{Perspective} & & \\
			& Distortion Scale & 0.4 \\
  			\textbf{Resize\&Crop} & & \\
			& Scale & $[0.7\dots1.0]$ \\
			\bottomrule

		\end{tabular}

	\end{center}
\end{table}

\section{Training Datasets}
\label{sec:apx_datasets}

This section gives more details on the data used during training of the described methods.
Overall, two different training dataset were used.
One taken with a robot-mounted camera used for comparing training with geometric and synthetic correspondence and for all results presented in section~\ref{sec:results}.
And another one with a fix-mounted camera used for the grasping experiment described in section~\ref{sec:experiment}.

\subsection{Dataset with robot-mounted camera}
 \label{app:dataset_rm}
\begin{figure}[h]
	\begin{subfigure}[b]{0.31\textwidth}
		\centering
		\includegraphics[width=\textwidth]{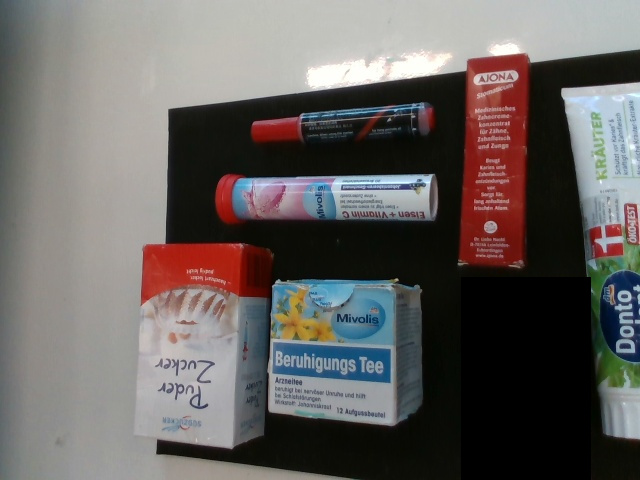}
	\end{subfigure}
	\hfill
	\begin{subfigure}[b]{0.31\textwidth}
		\centering
		\includegraphics[width=\textwidth]{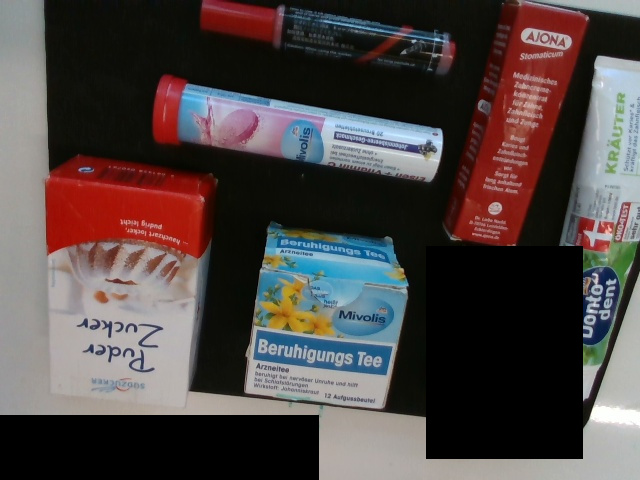}
	\end{subfigure}
	\hfill
	\begin{subfigure}[b]{0.31\textwidth}
		\centering
		\includegraphics[width=\textwidth]{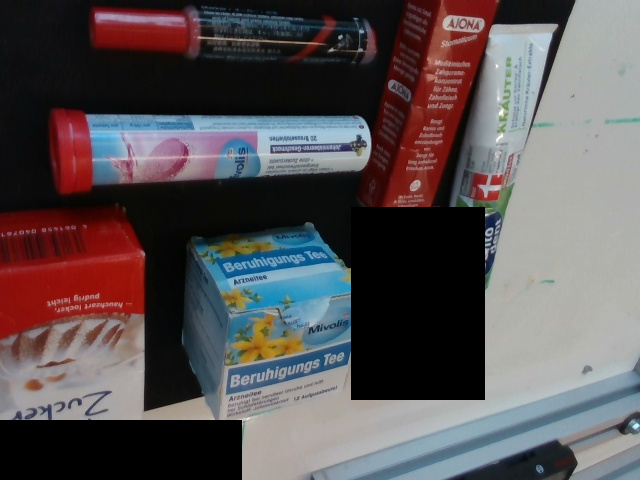}
	\end{subfigure}
\caption{Three example images of the mixed training dataset that was taken with a robot-mounted camera. Parts of the images where blacked-out to maintain anonymity of the authors.}
\label{fig:mixed_train_examples}
\end{figure}

 Fig.~\ref{fig:mixed_train_examples} shows example images from this dataset.
 We prepared seven objects in a static scene and recorded a stream of images with 30 frames per second, while the robot-mounted Realsense D435 camera moved in different perspectives around the scene.
 In total we took nine recordings like this with different configurations of the same objects.
 Six of those recordings were used for training, one for validation and one for testing.
 Each recording contains about 4400 images.

\subsection{Dataset for invariance tests}
\label{app:invariance_camera_movements}
\begin{figure}[h]
	\center
	\includegraphics[width=0.8\textwidth]{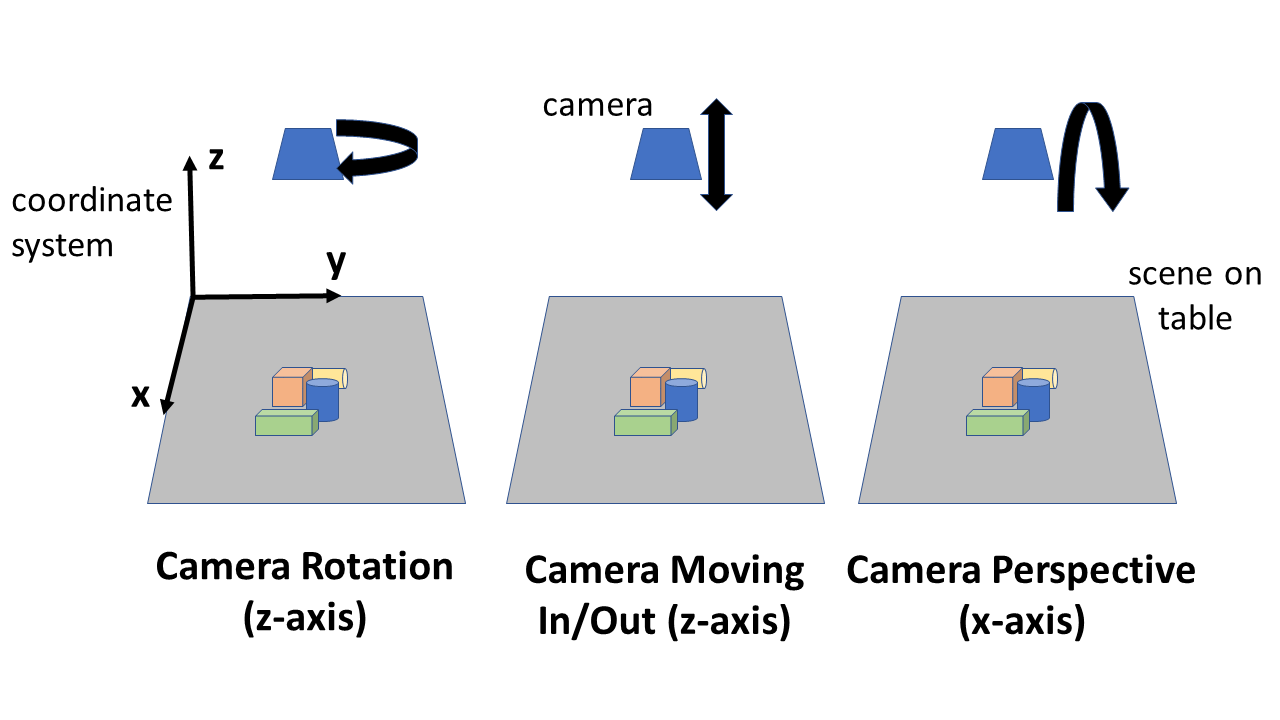}
	\caption{Camera transformations used for the invariance tests. Note thatt the origin of the coordinate system is located in the center of the scene on the table.}
	\label{fig:transformation_explanations}
\end{figure}

For the invariance tests reported in Sec.~\ref{sec:invariance_test} we took data with the same setup of the robot mounted camera and the same objects as reported above in Sec.~\ref{app:dataset_rm}, but with dedicated camera movements.
These camera transformations are visualized in Fig.~\ref{fig:transformation_explanations}.
The left shows the camera rotation ($z$-axis), where the location of the camera is fixed with the camera plane parallel to the table.
The camera is then rotated around the $z$-axis. 
In the middle we show the camera moving in/out along the $z$-axis, closer to and further away from the scene.
The $x$ and $y$-positions as well as the camera orientation are kept stable.
The last movement is the camera perspective movement.
Here, the camera is moved in $x$ direction on a half-sphere around the scene changing the orientation to keep focus on the center of the scene.
During this movement the distance of the camera to the center of the scene is fixed.

\subsection{Dataset with fixed-mounted camera}
\begin{figure}[h]
	\begin{subfigure}[b]{0.31\textwidth}
		\centering
		\includegraphics[width=\textwidth]{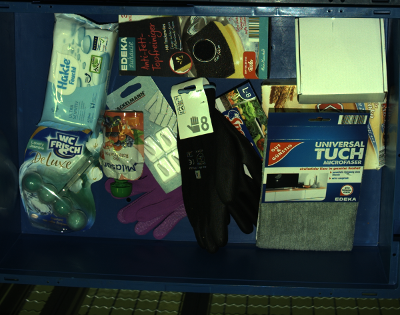}
	\end{subfigure}
	\hfill
	\begin{subfigure}[b]{0.31\textwidth}
		\centering
		\includegraphics[width=\textwidth]{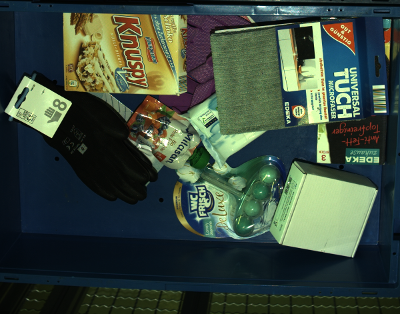}
	\end{subfigure}
	\hfill
	\begin{subfigure}[b]{0.31\textwidth}
		\centering
		\includegraphics[width=\textwidth]{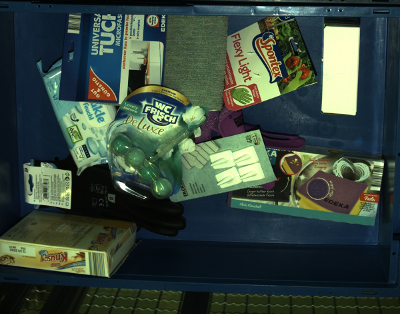}
	\end{subfigure}
\caption{Three example images of the training dataset for the grasping experiment.}
\label{fig:heaps_train_examples}
\end{figure}

Example images of this dataset are shown in Fig.~\ref{fig:heaps_train_examples}.
It was recorded with a fixed-mounted Zivid One+ camera.
We recorded 529 images of randomly shuffled heaps of the objects presented in section~\ref{sec:apx_grasping}.
Additional data was held back for validation and testing.

\section{Additional Results}
In the main section we provide results mostly in terms of the median pixel error and the 75\% quantile of pixel errors.
In the following section, we provide the main results with additional metrics.
Furthermore, we present results on the generalization capabilities on a small test set featuring unknown objects.

\subsection{Extended Main Results}
The results of Sec.~\ref{subsec:keypoint_tracking} are summarized in the Table~\ref{tab:full_kfold_table} with additional metrics.
	\begin{table}[ht]
		\begin{center}
			\caption{Pixel errors (mean, median, quantiles), and percentage of correct keypoints (PCK@k) metrics for geometric and synthetic correspondence training, evaluated on our kfold-cross validation dataset as used in Sec.~\ref{subsec:keypoint_tracking}.}
			\label{tab:full_kfold_table}
			\begin{tabular}{l|cc|ccc|ccccc}
				\toprule
    				 &       &        & \multicolumn{3}{c|}{Quantile} & \multicolumn{5}{c}{PCK@}\\
				Type &  Mean & Median & 75\% &  90\% & 95\% &  3 &  5 &  10 &  25 &  50 \\
				\midrule
				Geometric & 13.50 &    3.16 &      6.4 &    18.03 &    55.15 &   0.46 &   0.68 &    0.84 &    0.92 &    0.95 \\
				Synthetic & 21.49 &    5.10 &     11.0 &    50.25 &   103.25 &   0.29 &   0.49 &    0.73 &    0.85 &    0.90 \\
				\bottomrule
			\end{tabular}
		\end{center}
	\end{table}

\subsection{Results on Unknown Objects}
\begin{figure}[h]
	\begin{subfigure}[b]{0.45\textwidth}
		\centering
		\includegraphics[width=\textwidth]{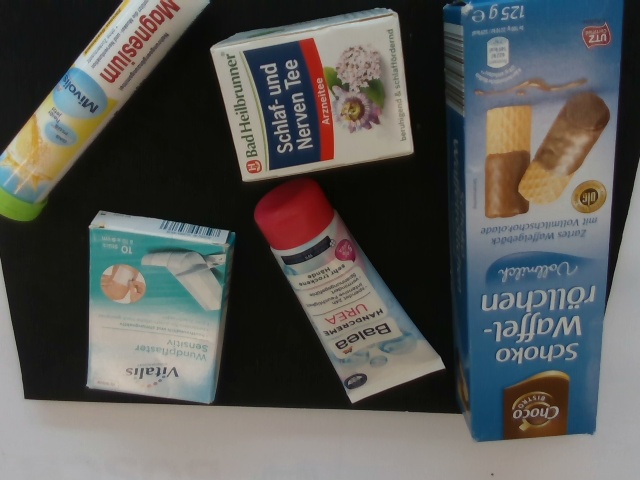}
	\end{subfigure}
	\hfill
	\begin{subfigure}[b]{0.45\textwidth}
		\centering
		\includegraphics[width=\textwidth]{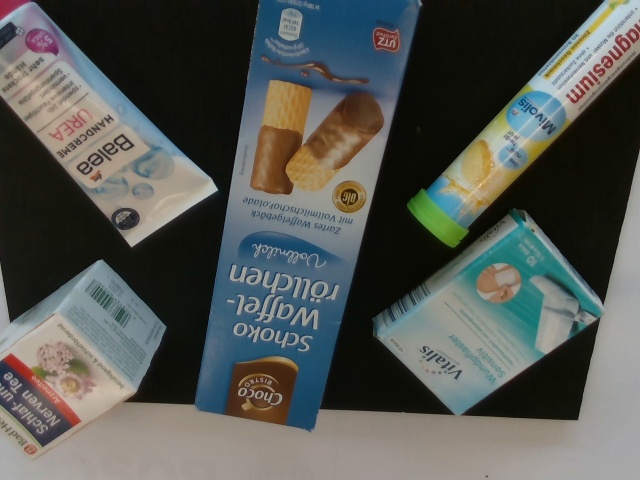}
	\end{subfigure}
	\caption{Novel object test set consisting of two new scenes, with 5 novel objects.}
	\label{fig:novel_objects}
\end{figure}

We further investigate the performance of our model on objects not seen during train time.
Both the geometric and synthetic correspondence models were trained on the same kfold dataset splits of the main section.
We test on five new objects, not previously seen during training and validation.
The objects and their arrangement in the two test scenes are shown in Figure~\ref{fig:novel_objects}.
We kept the training setup as in Sec.~\ref{subsec:keypoint_tracking}, although we note, that better generalization might be achieved with different configuration of hyper-parameters, the choice and amount of augmentation applied.
However, the overall trend is evident in the results compiled in Table~\ref{tab:unknown_objects}.

Both approaches, SV and GC, exhibit a loss in performance, especially the geometric correspondence training.
While the median changes only slightly, we find a large increase with respect to the 90\% and 95\% quantile for both methods.
Up to 25\% of the sampled keypoints are now mispredicted with an error nearly three times as high as before.

Using features from a purely pre-trained backbone, without further training, fails completely.
Training on a generic dataset, such as COCO, yields surprisingly good results, but still fails to work accurately.
For more details on the pre-trained and SV-COCO setup, see the ablation study in Section~\ref{sec:baseline_stoa}.

We note that, as we train only on a set of unordered RGB images, fine-tuning the model for additional new objects is as easy as adding a few new image taken of the novel objects.
Hence, despite limited generalization to completely new objects, the simple and efficient training of our proposal may effectively compensates for it.

\begin{table}[ht]
	\begin{center}
		\caption{Pixel errors (mean, median, quantiles), and percentage of correct keypoints (PCK@k) metrics for geometric and synthetic correspondence training, evaluated on two new scenes with 5 novel objects.}
	\label{tab:unknown_objects}
	\begin{tabular}{ccc|ccc|ccccc}
	\toprule
	     &        &        & \multicolumn{3}{c|}{Quantile} & \multicolumn{5}{c}{PCK@}\\
	Type &  Mean & Median & 75\% &  90\% & 95\% &  3 &  5 &  10 &  25 &  50 \\
	\midrule

	GC & 43.40 &    6.40 &   21.19 &  162.25 &  280.64 & 0.23 & 0.41 & 0.63 & 0.77 & 0.82 \\
	SV & 42.95 &    6.00 &   17.09 &  148.76 &  284.02 & 0.25 & 0.44 & 0.66 & 0.78 & 0.82 \\
	Pretrain Only & 49.96 &   22.47 &   42.01 &  112.70 &  260.41 & 0.03 & 0.08 & 0.21 & 0.55 & 0.80 \\
	SV-COCO & 42.35 &    4.12 &    9.43 &  165.06 &  302.76 & 0.36 & 0.58 & 0.76 & 0.82 & 0.85 \\
	\bottomrule
\end{tabular}
	\end{center}
\end{table}

\section{Ablation}
\subsection{Augmentation}
\label{sec:apx_ablation_augmentation}
To investigate the impact of each augmentation, we trained the synthetic approach with each combination and tested it on the same dataset as in Section~\ref{subsec:keypoint_tracking}.
The full results are listed in Table~\ref{tab:ablation_augmentation_full_results}.

\begin{table}[ht]
	\begin{center}
		\caption{Pixel error (mean, median, 75\% and 90\% quantile) for different combinations of augmentations for synthetic correspondence training. Abbreviations are as follows: \textbf{A}ffine, \textbf{P}erspective, \textbf{C}olor \textbf{J}itter, \textbf{R}e\textbf{s}ize \& \textbf{C}rop.}
		\label{tab:ablation_augmentation_full_results}
		\begin{tabular}{lrrrr}
			\toprule
			Combination &       Mean &     Median &    75\% Quantile & 90\% Quantile\\
			\midrule
			CJ + A + P + RSC &  17.18 &    5.00 &    10.05 &    37.48 \\
			A + P + RSC &  19.42 &    5.10 &    10.44 &    49.38 \\
			CJ + A + RSC &  19.62 &    4.47 &     9.49 &    55.63 \\
			A + RSC &  23.06 &    4.47 &    10.44 &    67.08 \\
			A + P &  36.64 &    8.60 &    21.02 &   104.05 \\
			CJ + A + P &  37.27 &    9.00 &    21.02 &   105.54 \\
			A &  52.46 &    6.71 &    49.65 &   181.03 \\
			CJ + A &  56.58 &    9.22 &    62.43 &   176.26 \\
			CJ + P + RSC &  97.10 &   20.81 &   168.44 &   296.19 \\
			P + RSC & 110.80 &   72.45 &   185.33 &   288.45 \\
			CJ + P & 144.76 &  118.43 &   226.37 &   342.49 \\
			P & 152.97 &  122.25 &   234.08 &   365.40 \\
			CJ + RSC & 176.97 &  164.47 &   255.64 &   345.12 \\
			RSC & 180.55 &  168.58 &   267.59 &   359.61 \\
			CJ & 227.43 &  211.21 &   316.31 &   412.00 \\
			\bottomrule
		\end{tabular}
	\end{center}
\end{table}

We see that affine transformations have a strong impact on the overall performance.
All other augmentations, even the combination of color jitter, perspective and resize+crops, performs considerably worse.
This result confirms that for our approach affine transformations are indeed essential to obtaining invariance to rotations with a CNN-based backbone network.
Generally, all combinations with affine augmentation further improve the models accuracy and robustness.
An exception is the combination of color jitter and affine, which seems to find a worse solution when combined.

We find that when using perspective distortion, the median typically seems to slightly decrease, while the mean, as well as the 75\% and 90\% quantiles improve considerably.
Hence, perspective distortions seem to play an important role in improving model robustness, but requires further investigation as to why the accuracy is negatively affected.

Color jitter seems to be the augmentation with the smallest impact.
However, we note that while we tested on different scenes, the lighting conditions are generally the same.
Hence, on datasets, and more importantly during model deployment, the impact of color jitter with respect to model robustness and reliability could be larger.

Lastly, we see that the combination of all augmentations ensure the learned descriptor is not overly focused on single type of invariance, but different kinds yielding the overall best result.

\subsection{Probability of Augmenting and Number of Augmented Frames}
In this experiment we vary two hyperparameters of our training: i) the chance that any given augmentation might be applied (independently drawn), ii) number of views that will be augmented.

\begin{figure}[h]
	\begin{subfigure}[b]{0.49\textwidth}
		\centering
		\includegraphics[width=\textwidth]{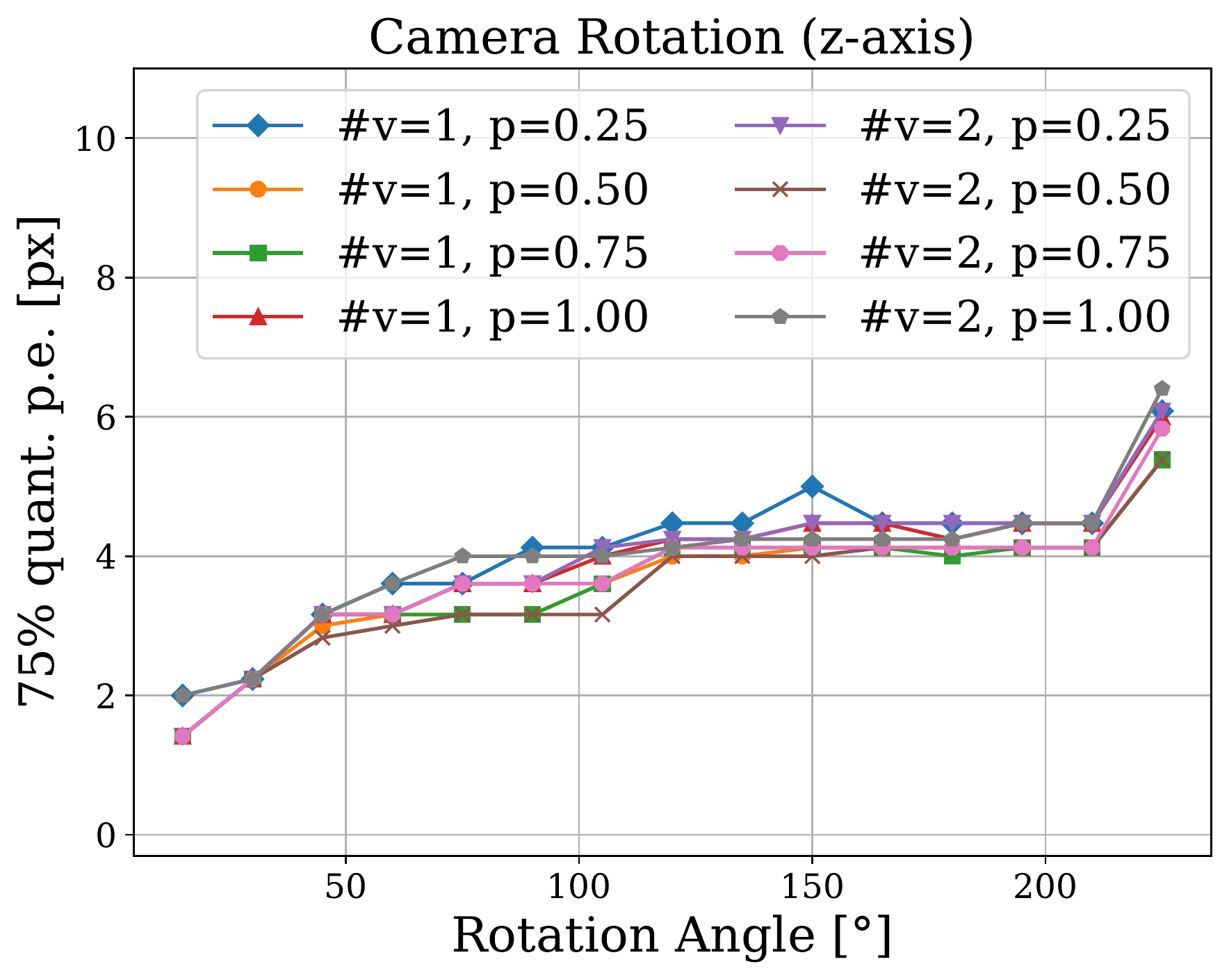}
	\end{subfigure}
	\hfill
	\begin{subfigure}[b]{0.49\textwidth}
		\centering
		\includegraphics[width=\textwidth]{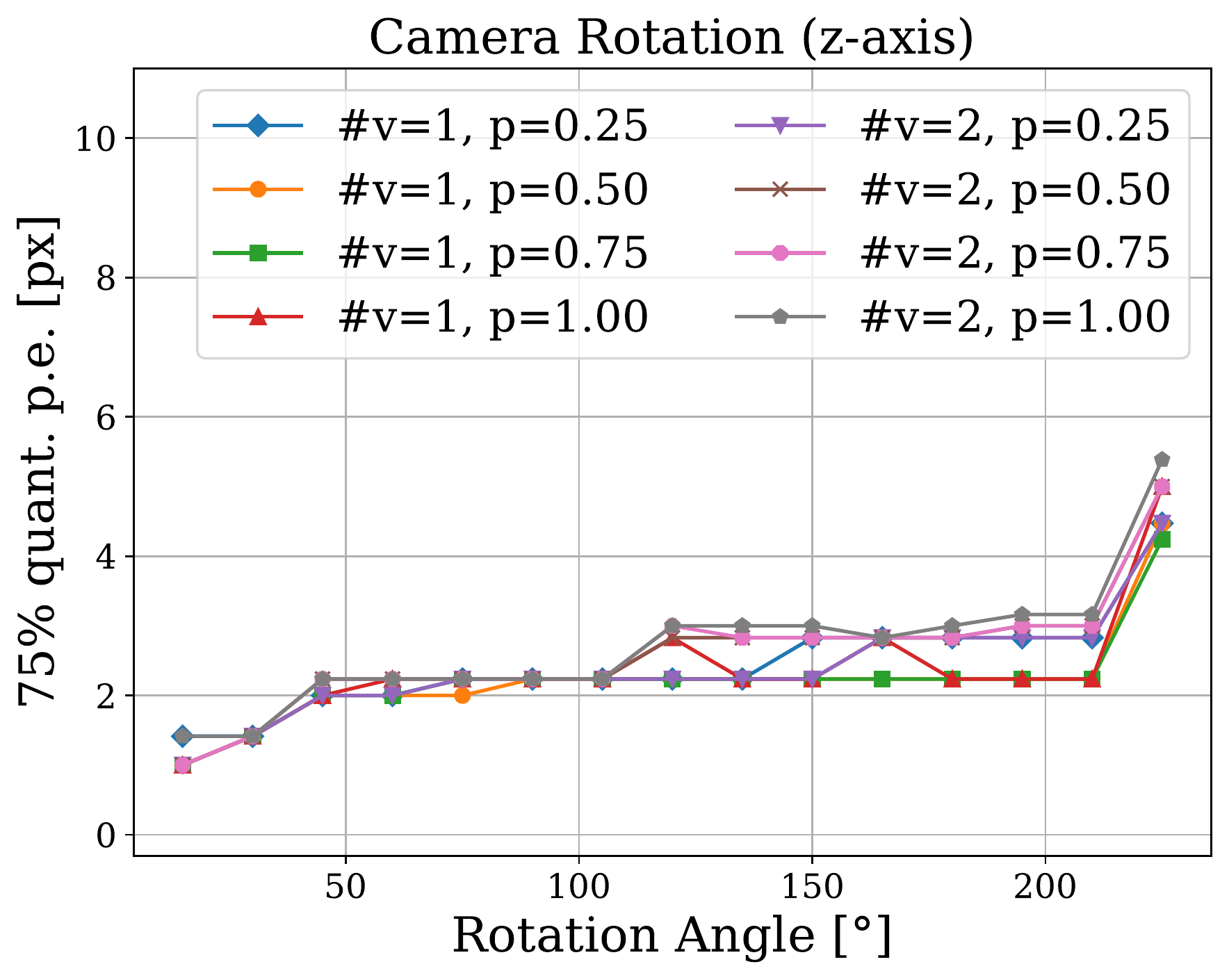}
	\end{subfigure}
	\\
	\begin{subfigure}[b]{0.49\textwidth}
	\centering
	\includegraphics[width=\textwidth]{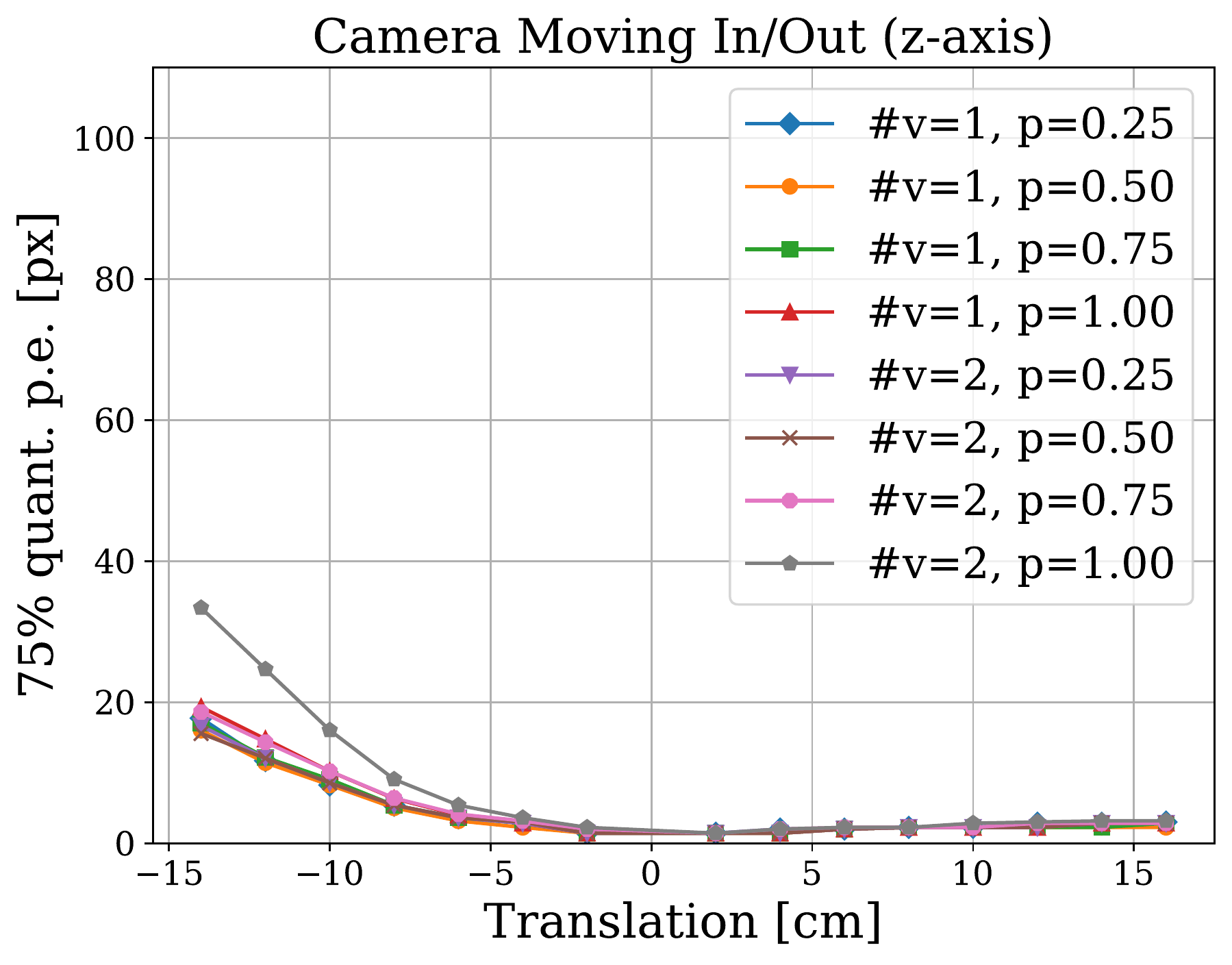}
	\end{subfigure}
	\hfill
	\begin{subfigure}[b]{0.49\textwidth}
		\centering
		\includegraphics[width=\textwidth]{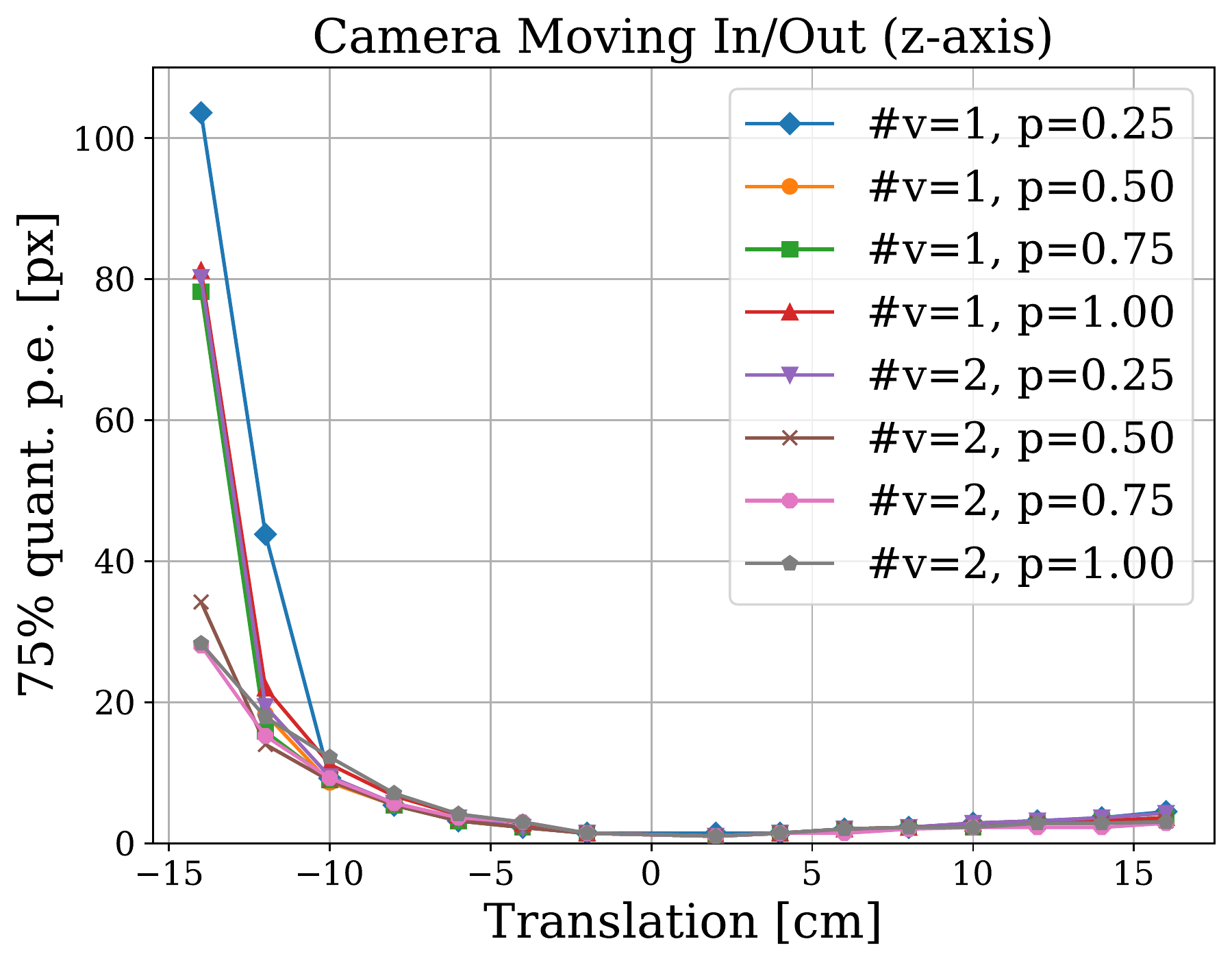}
	\end{subfigure}
	\\
		\begin{subfigure}[b]{0.49\textwidth}
		\centering
		\includegraphics[width=\textwidth]{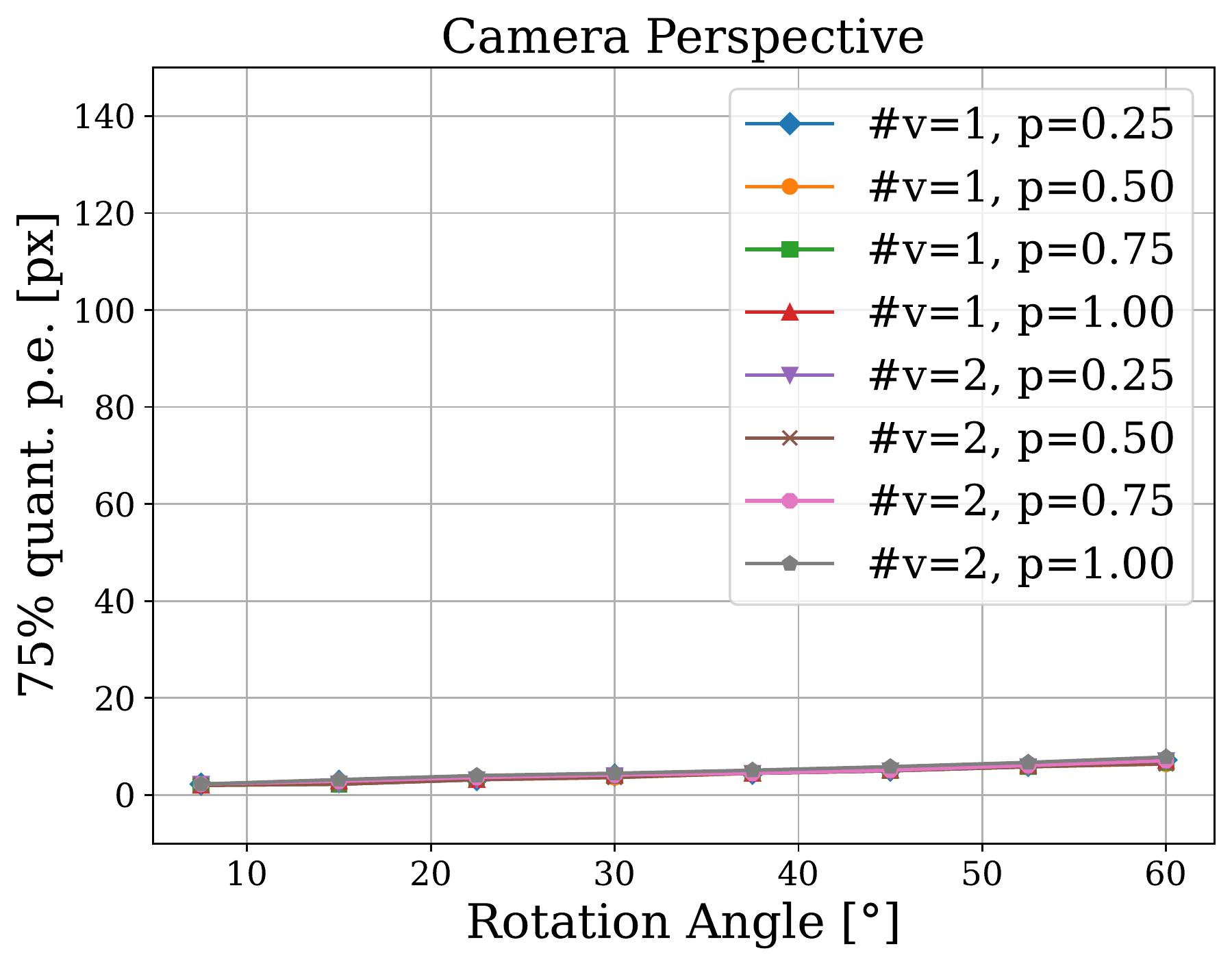}
		\caption{Trained using Geometric Correspondence}
	\end{subfigure}6
	\hfill
	\begin{subfigure}[b]{0.49\textwidth}
		\centering
		\includegraphics[width=\textwidth]{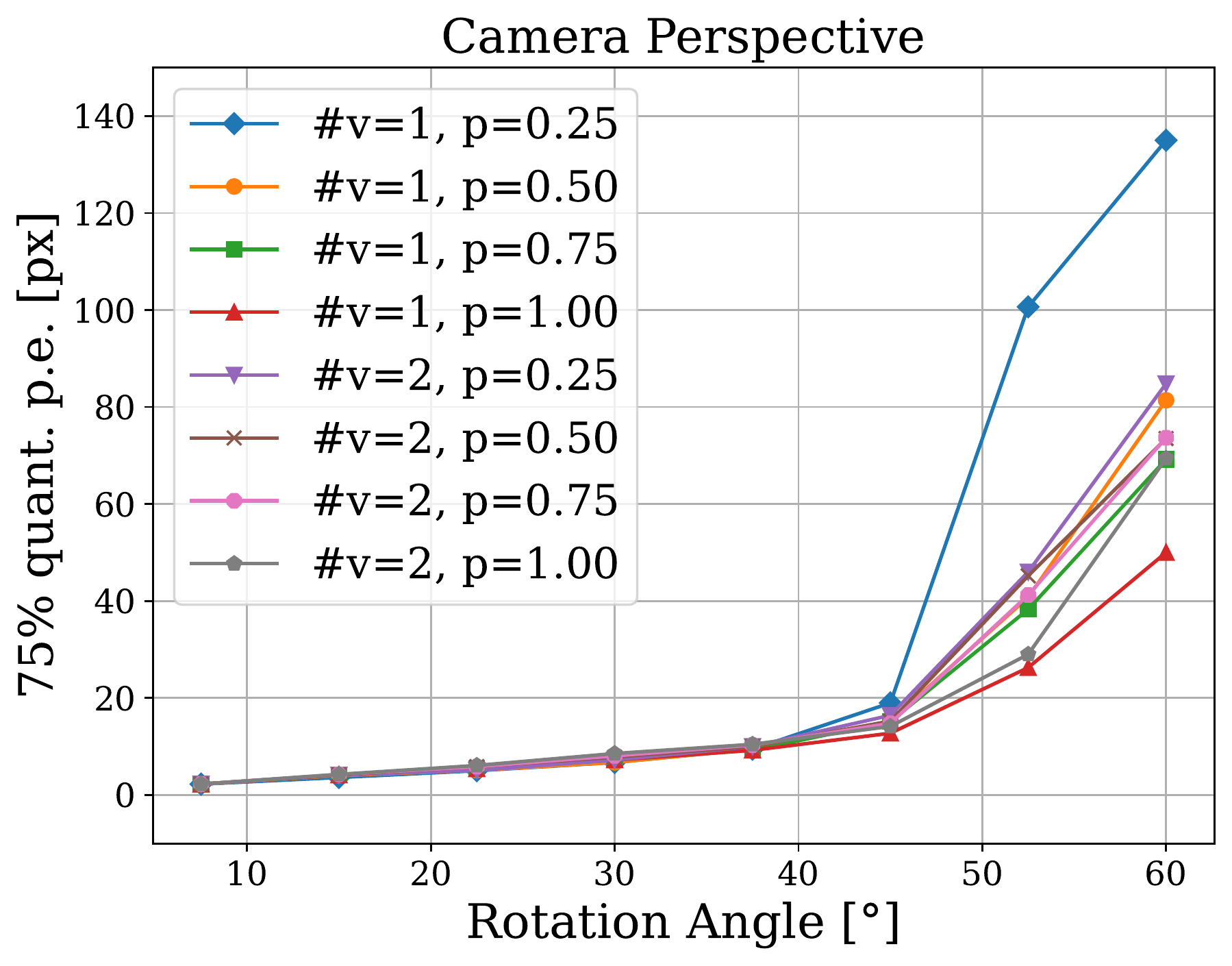}
		\caption{Trained using Synthetic Correspondence}
	\end{subfigure}
	\caption{75\% quantile pixel error for different combinations augmentation probabilities $p$, and number of augmented frames (\#v), evaluated with respect to different tasks: 
	(a) camera rotates around z-axis,
	(b) camera moving closer and further away from the objects,
	(c) camera moving on a sphere in x-direction around the objects, facing the objects.
	Note the scale of the y-axes.}
	\label{fig:invariance_sv_vs_dv}
\end{figure}

We evaluated each configuration on the invariance test dataset, with the results shown in Figure~\ref{fig:invariance_sv_vs_dv}.

We find that augmenting just one or both images, has limited impact for both geometric and synthetic correspondence training.
For geometric training we find that our setting, which is using 50\% probability per augmentation, yields similar results compared to augmenting just one view.
This was already observed in \cite{Adrian2022}.
We reconfirm, as reported by \cite{Adrian2022}, that augmenting more heavily, e.g., both frames with each augmentation at 100\%, has adverse effects on the performance of geometric correspondences trained networks.

In contrast, the synthetic correspondence training is most strongly impacted by the probability, less by the number of augmented images.
This is not surprising, as unlike the geometric training, augmentations are essential for the synthetic training, cf. Section~\ref{sec:apx_ablation_augmentation}.
Without any augmentations, both views are identical and the network will only learn a trivial solution.
Consequently, it is important to increase the chance, or guarantee in some ways, that at least one frame is augmented.
The difference between augmenting one or both views with high probability yields nearly the same results.
We note, that a more refined selection of differing probabilities per augmentation type would most likely yield even better results, rather than just one global parameter choice.

\subsection{Comparison to Baseline Methods and State-Of-The-Art Approaches}
\label{sec:baseline_stoa}
With this additional set of experiments we validate our assumption that domain specific data and explicit augmentations for perspective changes are crucial for a good performance on the target domain. For this, we compare our method to four different baseline methods:
\begin{enumerate}
	\item \textit{GC Specific}: Using data from different viewpoints (\cite{Schmidt2017, Florence2018, Adrian2022}), as already reported in the main paper.
	\item \textit{Pretrain only}: Using the features from a pretrained ResNet backbone (on ImageNet), without further fine-tuning. This serves as a na\"ive baseline. 
	\item \textit{SC COCO}: Using the method presented in this paper, but fine-tuned on COCO data instead of domain-specific data. 
	\item \textit{CATs}: A state-of-the-art keypoint matching algorithm among the top ranking methods on various keypoint matching datasets~\cite{Cho21}. We compare to the pretrained method on PF-Pascal as provided by the authors.
\end{enumerate}

\begin{figure}[t] 
	\centering
	\begin{subfigure}{0.32\textwidth}
		\includegraphics[width=1.0\textwidth]{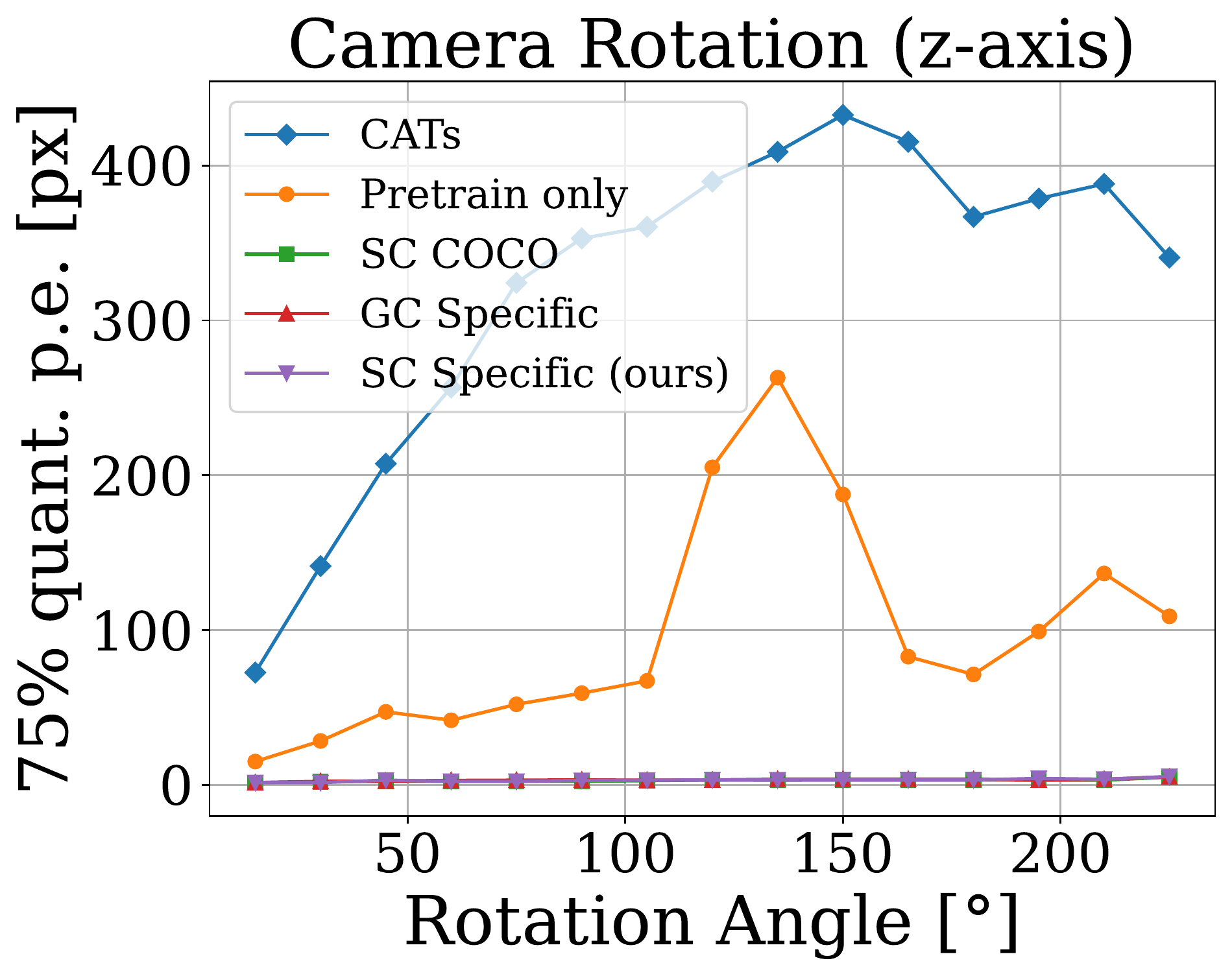}
		\caption{}
	\end{subfigure}
	\hfill
	\begin{subfigure}{0.32\textwidth}
		\includegraphics[width=1.0\textwidth]{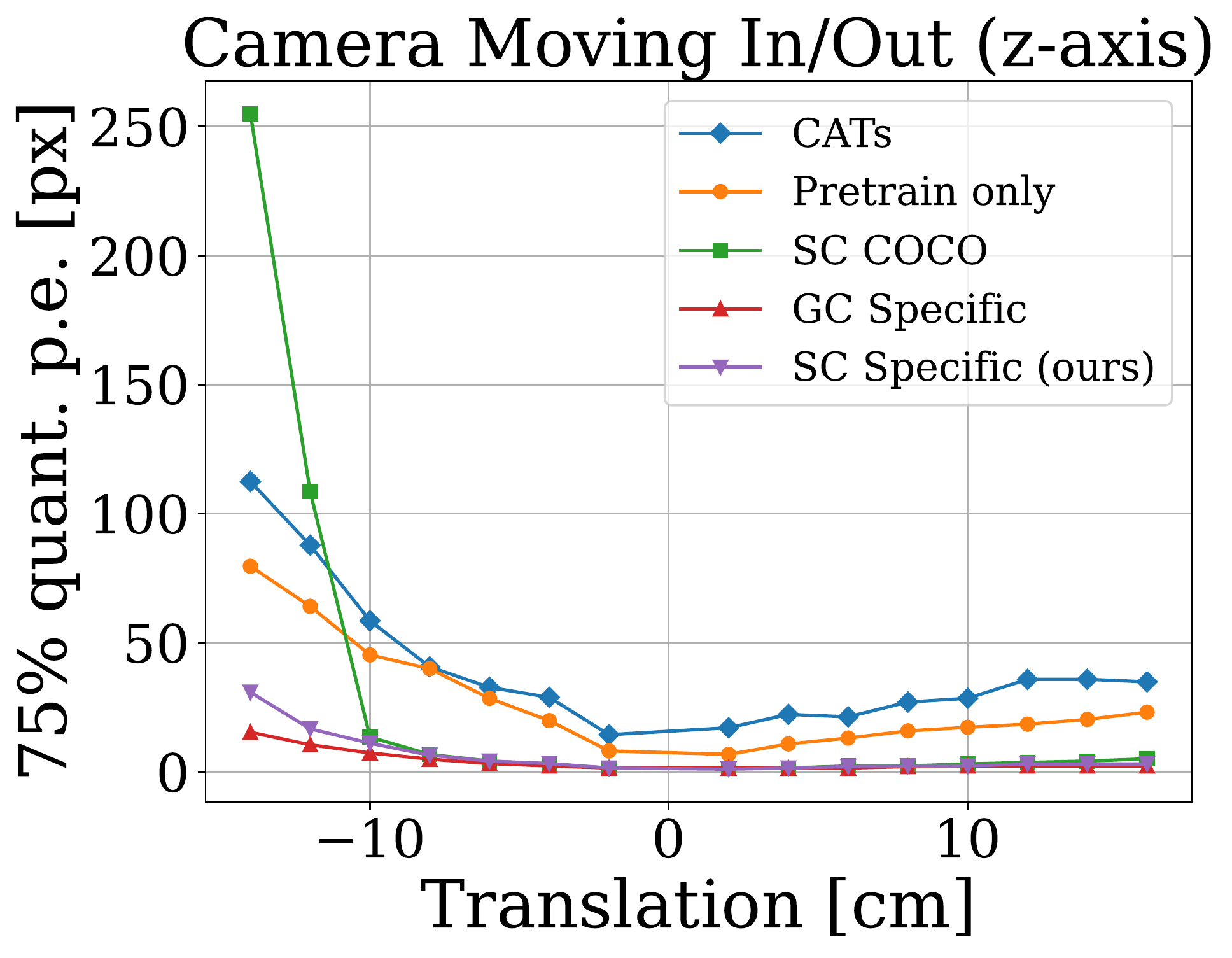}
		\caption{}
	\end{subfigure}
	\hfill
	\begin{subfigure}{0.32\textwidth}
		\includegraphics[width=1.0\textwidth]{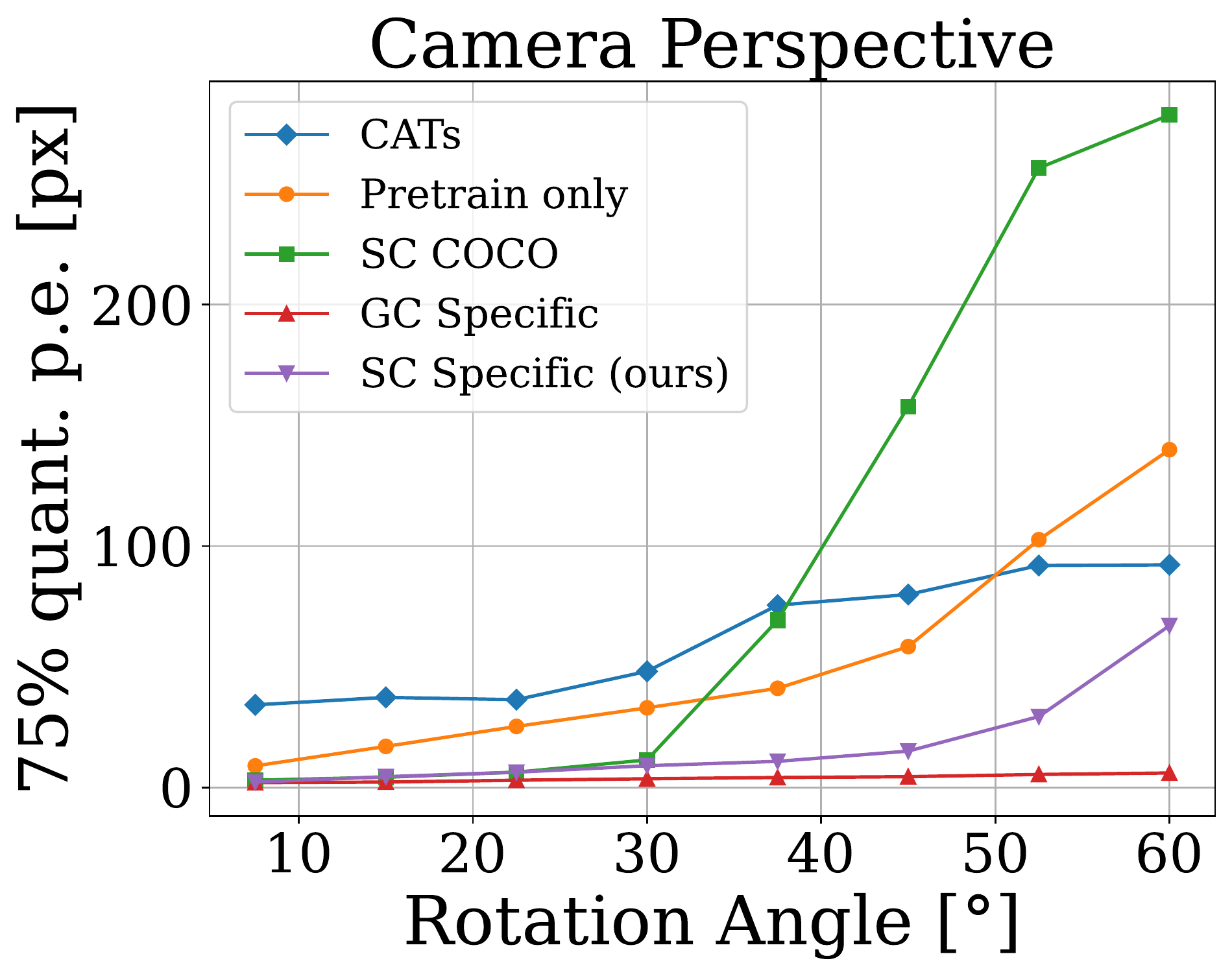}
		\caption{}
	\end{subfigure}
	\caption{75\% quantile pixel error for different network and data configurations.
		Networks named \textit{SC} are trained using our proposed synthetic correspondence setup, \textit{GC} using geometric correspondences, and lastly, as sanity check, raw features extracted from our ImageNet-pretrained ResNet backbone are named \textit{Pretrain only}.
		\textit{SC COCO} was trained using the COCO dataset, while networks named \textit{Specific} were trained using our dataset.
		We compare with respect to three view invariance tasks: 
		(a) camera rotates around z-axis,
		(b) camera moving closer and further away from the objects,
		(c) camera moving on a sphere in x-direction around the objects, facing the objects.
		Note the scale of the y-axes.}
	\label{fig:coco_comparison}
\end{figure}

We evaluated all methods on our view-invariance test dataset, as described in~\ref{sec:invariance_test}. Figure~\ref{fig:coco_comparison} shows the results.

Not surprisingly, the raw pretrain-only features, exhibit little to no rotational invariance, and generally lack view-invariance on other tests.

\textit{SC COCO} does perform better than the pretrained-only method, especially for smaller transformation angles. However, it seems to not generalize well to larger transformation angles in our invariance tests. This indicates that in-distribution training data is important for the accuracy we need for the robotics use-case.

The keypoint matching method \textit{CATs} achieves very impressive results for semantic keypoint matching, where e.g., the tip of a dog’s nose will be matched to a completely different dog’s nose in a second image. Surprisingly, the method does not outperform the \textit{Pretrain only} baseline on our test dataset. We account this result to the following: (a) PF-Pascal has a limited number of classes and the pretrained network overfits to those (b) The goal of \textit{CATs} (and similar approaches) is semantic keypoint matching. In this goal it achieves very impressive results. However, the goal of these methods are not the very accurate matching of geometric points on target objects, which we need for robotic grasping (see the reported mean pixel error in the original \textit{CATs} paper).

The above experiments support our claim that that our proposed training schema with the choice of augmentations and loss together with in-distribution training data containing scenes of the target objects, plays an important role to achieve sufficiently high accuracy for robotic grasping applications.

\section{Grasping Experiment}
\label{sec:apx_grasping}
\subsection{Objects}

\begin{figure}[h]
	\begin{subfigure}[b]{0.14\textheight}
		\centering
		\includegraphics[width=\textwidth]{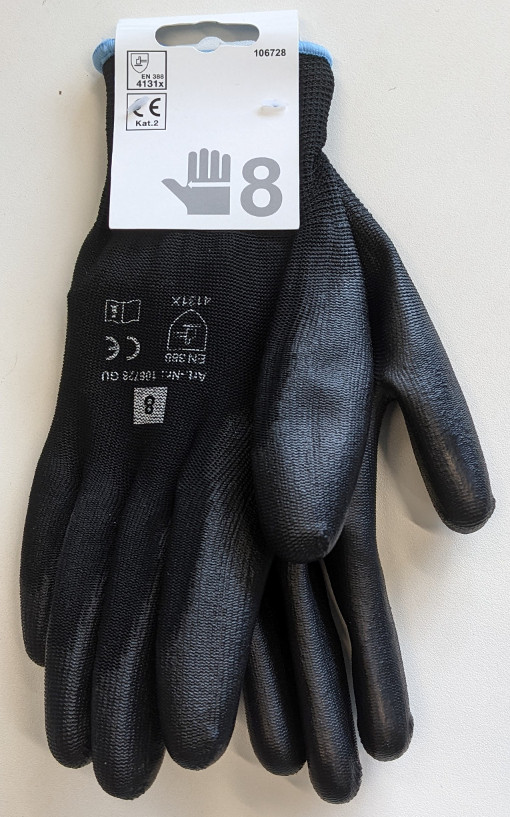}
     	\caption{}
		\label{fig:objects_gloves_black}
	\end{subfigure}
\begin{subfigure}[b]{0.11\textheight}
	\centering
	\includegraphics[width=\textwidth]{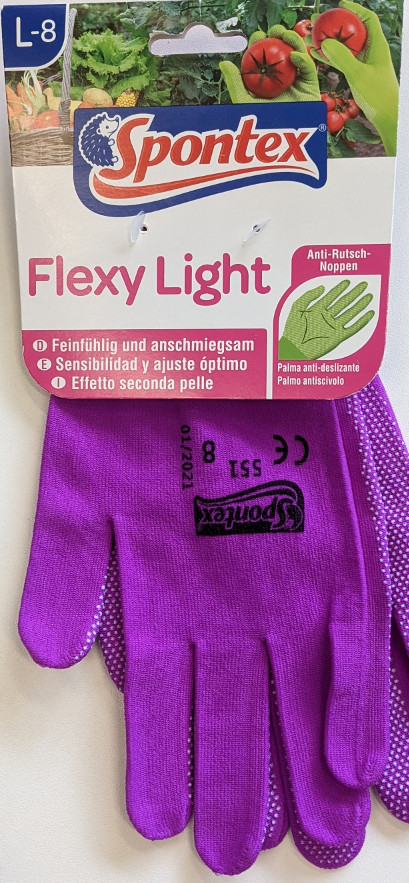}
	\caption{}
    \label{fig:objects_gloves_pink}
\end{subfigure}
\begin{subfigure}[b]{0.11\textheight}
	\centering
	\includegraphics[width=\textwidth]{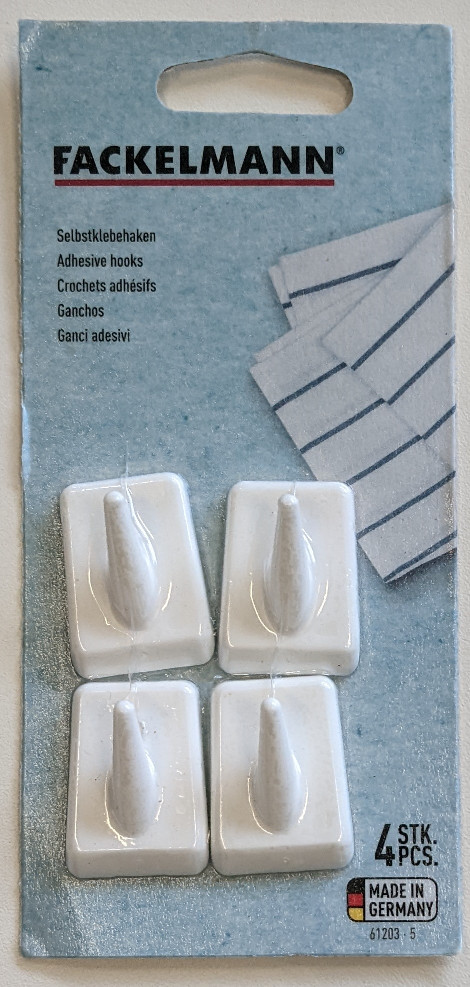}
	\caption{}	
    \label{fig:objects_hangers}
\end{subfigure}
\begin{subfigure}[b]{0.11\textheight}
	\centering
	\includegraphics[width=\textwidth]{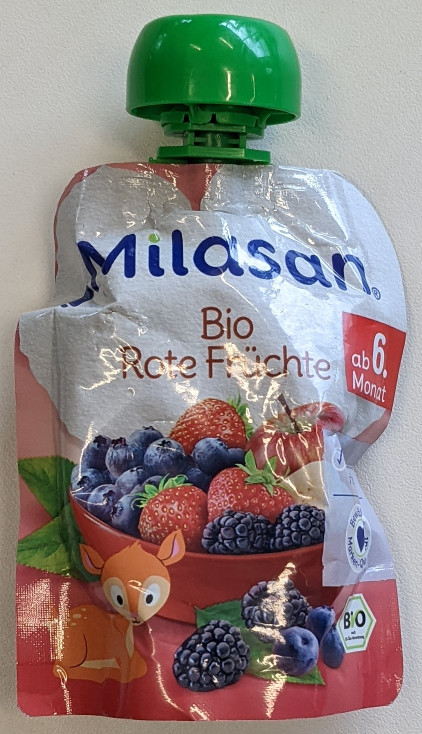}
	\caption{}	
    \label{fig:objects_milasan}	
\end{subfigure}
\begin{subfigure}[b]{0.11\textheight}
	\centering
	\includegraphics[width=\textwidth]{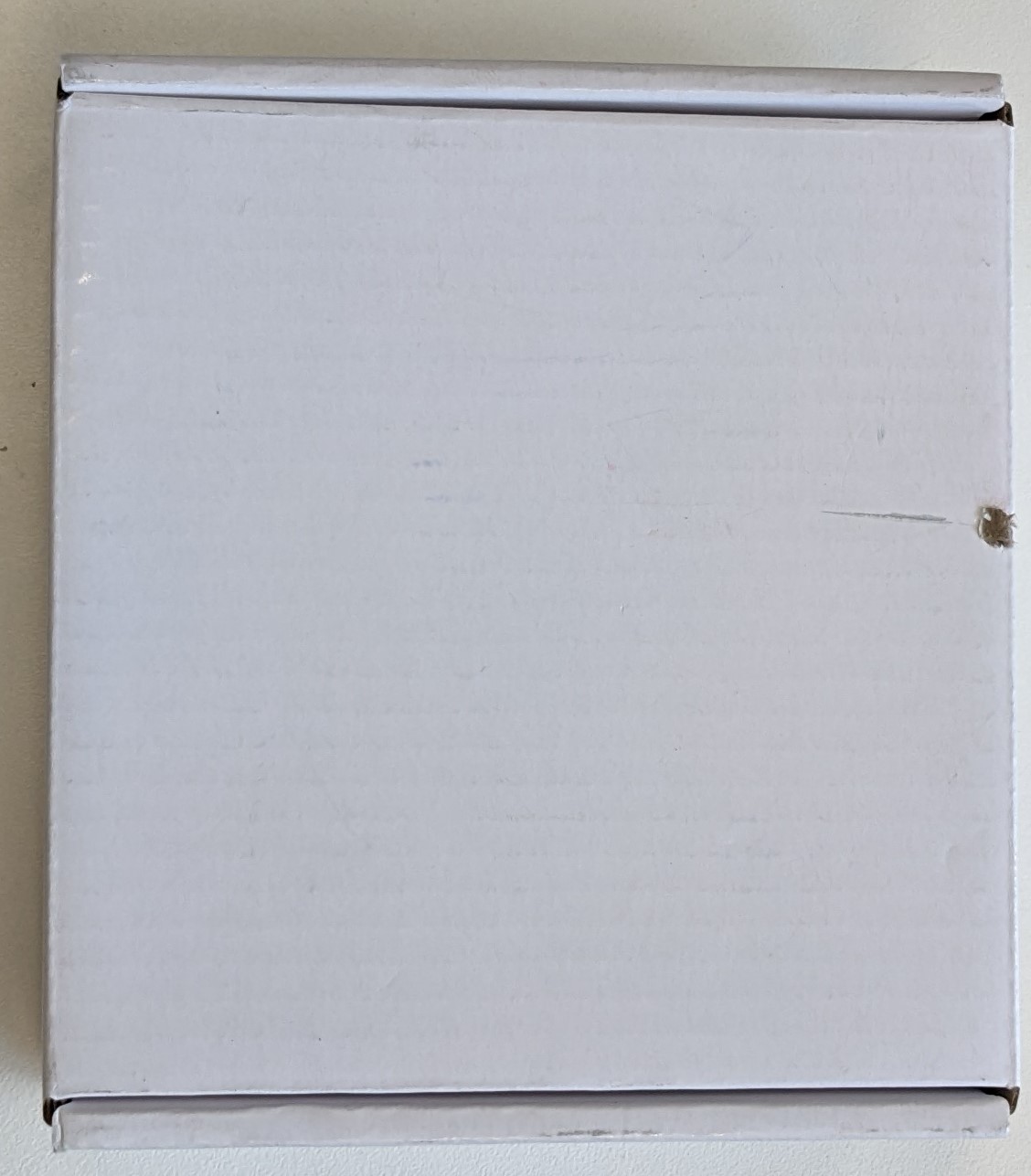}
	\caption{}
    \label{fig:objects_whitebox}
\end{subfigure}
\\
\begin{subfigure}[b]{0.12\textheight}
	\centering
	\includegraphics[width=\textwidth]{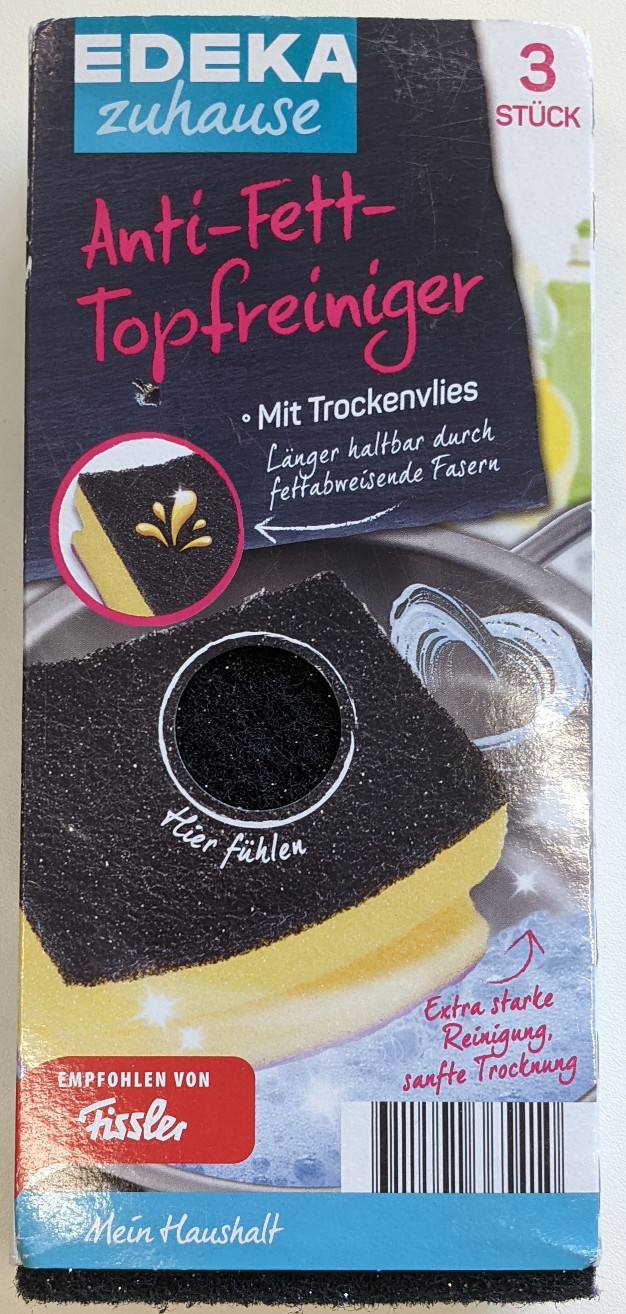}
	\caption{}
    \label{fig:objects_sponge}	
\end{subfigure}
\begin{subfigure}[b]{0.15\textheight}
	\centering
	\includegraphics[width=\textwidth]{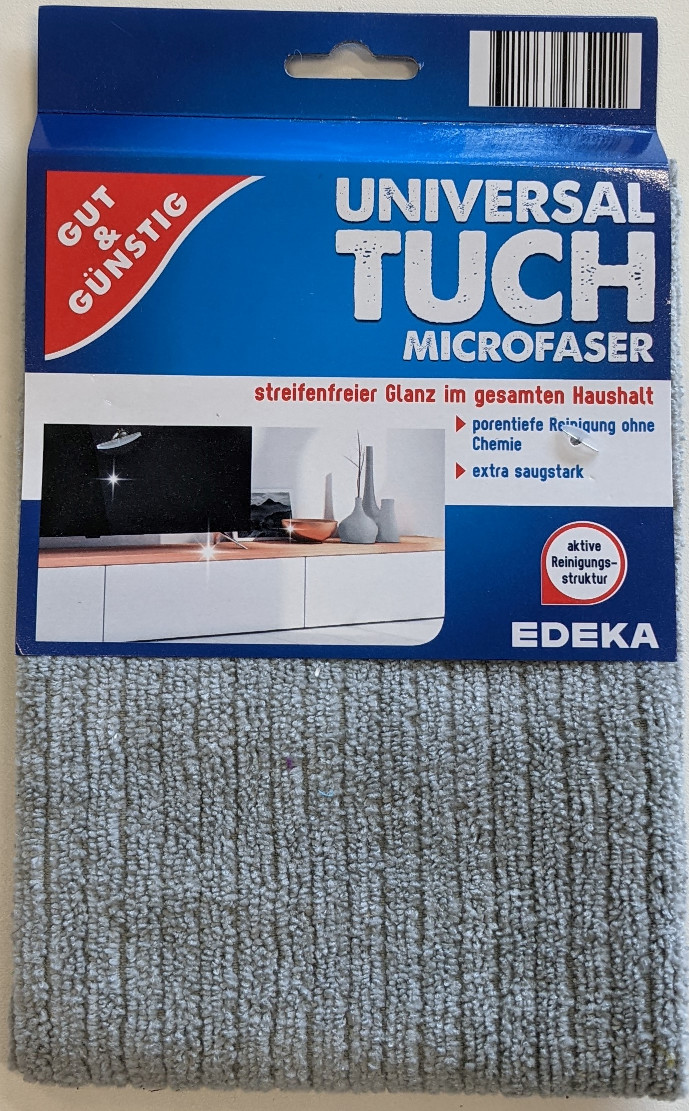}
	\caption{}
    \label{fig:objects_tuch}		
\end{subfigure}
\begin{subfigure}[b]{0.16\textheight}
	\centering
	\includegraphics[width=\textwidth]{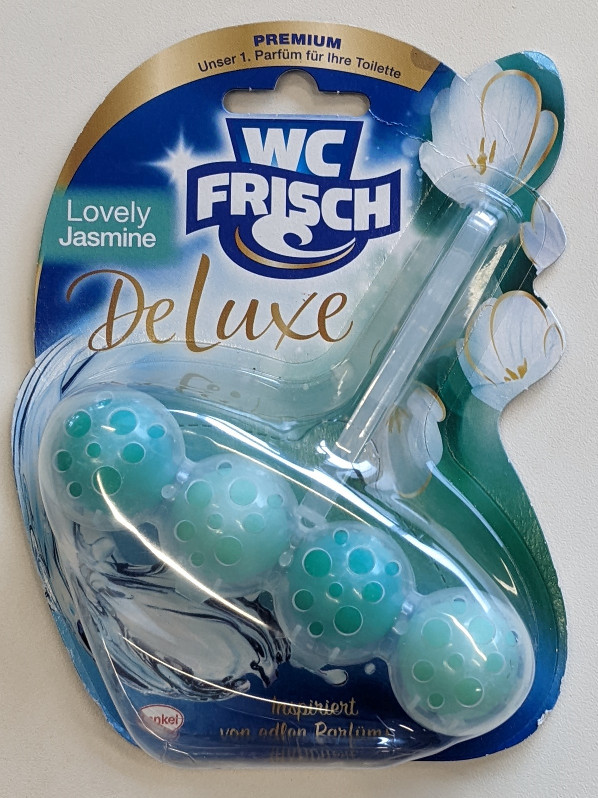}
	\caption{}
    \label{fig:objects_wcfrish}	
\end{subfigure}
\begin{subfigure}[b]{0.13\textheight}
	\centering
	\includegraphics[width=\textwidth]{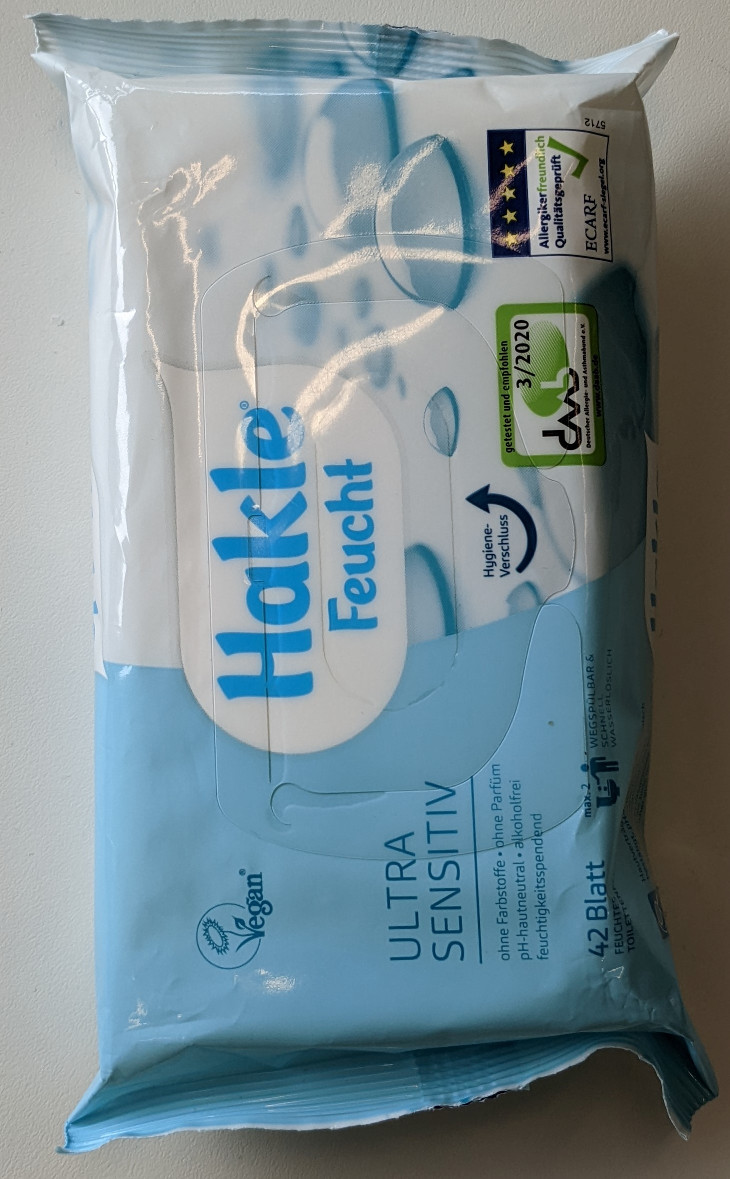}
	\caption{}
    \label{fig:objects_wetwipe}
\end{subfigure}
\caption{
	We used nine different objects in the grasping experiment.
	Some of these objects are prone to fail when planning the grasp based on 3D geometry.
	In other cases a suction grasp may damage parts of the objects.
	Therefore, using grasp preferences both improves grasp success and avoids damaging object packaging.}
\label{fig:objects}
\end{figure}

In Fig.~\ref{fig:objects} we show the objects we used in the grasping experiment.
Each object is either challenging to grasp with a suction gripper while relying only on depth images and geometrical features, or successful grasps may damage the objects.
Therefore, we wish to rely on human annotated grasp preferences to avoid damage and improve success chances.
Fig.~\ref{fig:objects_gloves_black} and Fig.~\ref{fig:objects_gloves_pink} show gloves which are only graspable on the paper label, which is also the preferred grasp location.
Additionally, a cutout at the top and plastic strips in the middle of the paper label make these objects challenging to grasp.
The depth camera may not recognize the small bumps in the depth image for the hangers in Fig.~\ref{fig:objects_hangers}, therefore we would like to enforce grasping on the paper label.
The non-rigid object in Fig.~\ref{fig:objects_milasan} is not particularly hard to grasp, but we wish to improve success chance by grasping in the middle of the object.
Similarly, the white box in Fig.~\ref{fig:objects_whitebox} is not challenging to grasp, but grasping in the middle improves the chance of success.
The dense descriptor representation allows to accurately locate the center for such textureless objects.
The sponge in Fig.~\ref{fig:objects_sponge} has a cutout in the middle of the packaging where suction grasps will fail, therefore we prefer to grasp towards the top, or bottom.
As the towel in Fig.~\ref{fig:objects_tuch} is clearly visible in the depth image, suction grasps will often fail on the towel, but not on the label. 
The plastic cover in Fig.~\ref{fig:objects_wcfrish} does not show up on depth images, therefore we prefer to focus the grasp towards the top of the packaging.
Finally, the wet wipe in Fig.~\ref{fig:objects_wetwipe} is not difficult to grasp, but we wish to avoid grasping by the package opening, which might be damaged when using a suction gripper.

\subsection{Generating Grasp Preference Heatmaps}

In this section we detail the computation steps of the grasp preference heatmap.
The experiment consists of an offline grasp preference annotation phase, and an online autonomous operation phase performing the grasps in the bin.

\textbf{Annotation phase.} First an RGB image $I$ showing the objects in the bin is presented.
Then the human clicks at pixel locations $\{k_i^j\}$ corresponding to preferred grasp locations.
The descriptor values $d_j=f(I;\theta)(k_i^j)$ at these pixel locations are then stored into a keypoint database $\mathcal{D}=\{d_j\}$.
See Fig.~\ref{fig:grasp_illustration_rgb} for an example image of the objects in the bin in a random configuration. 

\textbf{Autonomous operation phase.} During autonomous operation, the latest RGB image $I$ taken of the bin is evaluated with the trained network resulting in the descriptor image $I_d = f(I;\theta)$.
Then, keypoint heatmaps are generated from the database with $h_j(u, v) = \exp(-\dist{I_d, d_j} / \eta),~\forall d_j \in \mathcal{D}$, with $\eta$ as a temperature parameter that controls the width of the heatmap.
Finally, the individual keypoint heatmaps are fused into a single heatmap function $h(u,v) = \sum_j h_j(u,v) / H$, with $H$ as a normalization constant (see Fig.~\ref{fig:grasp_illustration_heatmaps} for an illustration).

\end{document}